%% file: main.tex
\documentclass[10pt,twocolumn,letterpaper]{article}

\usepackage{3dv}
\usepackage{times}
\usepackage{epsfig}
\usepackage{graphicx}
\usepackage{amsmath}
\usepackage{amssymb}

\usepackage{color}
\newcommand\todo[1]{}
\usepackage{tabularx}

\usepackage[pagebackref=true,breaklinks=true,letterpaper=true,colorlinks,bookmarks=false]{hyperref}

\threedvfinalcopy 

\def\threedvPaperID{129} 

\ifthreedvfinal\fi
\begin{document}

\title{Deep SVBRDF Estimation on Real Materials}

\author{Louis-Philippe Asselin \quad Denis Laurendeau \quad Jean-François Lalonde\\
Université Laval\\
{\tt\small louis-philippe.asselin.2@ulaval.ca, \{denis.laurendeau, jflalonde\}@gel.ulaval.ca}}

\maketitle

\begin{abstract}
    
Recent work has demonstrated that deep learning approaches can successfully be used to recover accurate estimates of the spatially-varying BRDF (SVBRDF) of a surface from as little as a single image. Closer inspection reveals, however, that most approaches in the literature are trained purely on synthetic data, which, while diverse and realistic, is often not representative of the richness of the real world. In this paper, we show that training such networks exclusively on synthetic data is insufficient to achieve adequate results when tested on real data. Our analysis leverages a new dataset of real materials obtained with a novel portable multi-light capture apparatus. Through an extensive series of experiments and with the use of a novel deep learning architecture, we explore two strategies for improving results on real data: finetuning, and a per-material optimization procedure. We show that adapting network weights to real data is of critical importance, resulting in an approach which significantly outperforms previous methods for SVBRDF estimation on real materials. Dataset and code are available at \urlstyle{same}\url{https://lvsn.github.io/real-svbrdf}.


\end{abstract}


\input{content/introduction}

\section{Related work}
\input{content/related_work}

\input{content/capture}

\input{content/method}

\def\myfigtype{ssim} 
\input{content/experiments}

\input{content/discussion}

\section*{Acknowledgements}

This work was supported by the REPARTI Strategic Network and the NSERC/Creaform Industrial Research Chair on 3D Scanning: CREATION 3D. We thank Charles Asselin for helping with data capture, Pierre Robitaille for electronics, Yannick Hold-Geoffroy for his invaluable proofreading skills, and Nvidia with the donation of GPUs.


{\small
\bibliographystyle{ieee}
\bibliography{egbib}
}

\end{document}

%% file: content/introduction.tex
\section{Introduction}

Photometric stereo, or the process of estimating surface normals from images under different illumination, originates decades ago~\cite{woodham1980photometric}. It has since been extended to recover other surface properties such as its bidirectional reflectance distribution function, or BRDF~\cite{alldrin2008photometric,goldman2009shape}. Unfortunately, the recovery of rich and expressive BRDFs is done at the expense of complicated capture setups, requiring the acquisition of hundreds of carefully-calibrated images. 

Recently, the advent of deep learning has enabled the development of techniques proposing to estimate spatially-varying BRDFs (SVBRDFs---different BRDF parameters for each pixel on a surface patch) from as little as a \emph{single} image~\cite{Li:2017:MSA,Li_2018,Valentin_2018,Gao_2019}. While not as accurate as physics-based techniques relying upon large number of images, approaches based on deep learning have democratized BRDF acquisition since they have shown that a cellphone camera and its flash are sufficient to capture real-world materials and reproduce them in a virtual environment. 

However, a closer inspection of these technique reveals that an important---and possibly quite limiting---assumption is made. Indeed, because of the difficulty in acquiring ground truth SVBRDF data, most recent papers on the topic train their approaches exclusively on synthetically-generated data and hypothesize that the trained models will generalize well on real data. Some approaches do propose a fine-tuning step on real photos (for example, \cite{Gao_2019,Deschaintre2020}), but it is still unclear how they perform quantitatively since no systematic evaluation on real data has been performed. 

This paper aims to fill that gap by providing the first evaluation of deep learning SVBRDF estimation techniques on real materials. To do so, we designed a portable and convenient multi-light capture apparatus which we use to acquire images of a wide variety of real materials from multiple lighting directions. We also present a novel network architecture that can be configured in various ways to reproduce, and improve upon, most previous works. Armed with these, we proceed to demonstrate that, as in many other problems (e.g., \cite{ganin2015unsupervised, chen2019learning, maximov2020focus}), there exists a significant gap between synthetic and real data and that training solely on synthetic data does not yield acceptable SVBRDF results on real materials. Therefore, additional steps must be performed on real data to bridge this gap. We thus experiment with two choices: fine-tuning the network on a small set of real examples which is beneficial for time-critical applications; and optimizing the network weights on a single exemplar at a time when higher quality is desired. 

In summary, the main contributions of the paper are the following. First and foremost, it presents the first systematic evaluation of deep learning SVBRDF estimation techniques on real materials. Second, it introduces a novel dataset of real materials captured under different lighting conditions that can be used to quantify performance in SVBRDF estimation through a rendering loss. Third, it presents a novel deep learning architecture, inspired by StyleGANv2~\cite{Karras2019stylegan2}, for estimating SVBRDFs from one or many input images captured by a portable multi-light capture apparatus.

%% file: content/related_work.tex

\paragraph{Calibration and complex capture setups}
At one end of the spectrum of BRDF capture systems, gonioreflectometers~\cite{Foo97agonioreflectometer} or other dedicated systems~\cite{meka2019deep} achieve high accuracy material and shape recovery. However, these systems are typically complex and require a prohibitive amount of time and processing power to execute. 
At the other end of the spectrum reside simpler methods that use common devices such as handheld cellphones to capture a single image~\cite{Li_2018, Valentin_2018, Li2019_shape}, with or without a flash~\cite{aittala2015two}, or multiple images~\cite{Valentin_2019, Gao_2019, Boss_2020_shape}. 
When using a renderer to evaluate the estimated SVBRDF, knowing the positions of light sources and camera is required to compare with real images. Recent work~\cite{Gao_2019} attempts to mitigate this issue by estimating the rendering configuration along with the SVBRDF.
In line with Schmitt~\etal~\cite{Schmitt2020CVPR}, our \emph{portable} RGB-D capture system proposes a hybrid solution that allows for fast and accurate image acquisition. We employ this system to capture a new dataset of real-world materials, and compare various SVBRDF estimation strategies on this data. 

\paragraph{Reflectance models} 
Estimation of simple surface reflectance models, such as lambertian, was traditionally investigated using Photometric Stereo (PS), which focuses on estimating surface normals and albedo~\cite{woodham1980photometric} from a static camera with a moving light source. Since then, optimization- or deep learning-based PS approaches were proposed to extend to specular or spatially-varying BRDFs~\cite{alldrin2008photometric,goldman2009shape, shi2016benchmark, ikehata2018cnn, PS-FCN,PS-SPLINE-Net, Chen_2019, enomoto2020photometric}. More complex material models such as~\cite{GGX, DisneyBRDF, UnrealBRDF} were also proposed to better capture the rich diversity of real materials. Recent minimal PS approaches~\cite{li2019learning}  need only six input images, but focus on surface normals only and do not recover full SVBRDFs.

\paragraph{Image-based SVBRDF estimation}
Due to the complexity of these reflectance models, recent work leverages deep learning to robustly estimate material parameters from images. Those data-driven methods typically require a large dataset for proper training.
Due to the difficulty of acquiring labeled SVBRDF samples, especially for real material datasets, training is generally performed on synthetic renders using SVBRDFs generated by artists. Performance evaluation is sometimes based on a small collection of real images often captured by uncalibrated setups~\cite{aittala2015two, Li_2018,Valentin_2018, Valentin_2019, Gao_2019, Schmitt2020CVPR, Deschaintre2020}.
Other methods estimate the SVBRDF alongside the geometry of the object~\cite{Li2019_shape, li2019inverse, Schmitt2020CVPR}.

\paragraph{Refinement}
It has been shown that a post-training refinement step greatly improves SVBRDF estimation~\cite{Li_2018, Gao_2019}, especially for deep learning methods that do not use iterative or cascade networks~\cite{Li2019_shape}. Li~\etal~\cite{Li_2018} enhance their CNN prediction by using DCRFs. Gao~\etal~\cite{Gao_2019} train an auto-encoder to learn the distribution of plausible SVBRDFs and optimize on the latent space of the encoder. Similarly, work on neural-based rendering~\cite{Tewari_2020} has recently been used as an extra supervision or refinement step in addition to classical rendering relying on SVBRDF maps. In concurrent work, Deschaintre~\etal~\cite{Deschaintre2020} estimate large SVBRDF maps by fine-tuning a model for each material and by using various patches of the material.

%% file: content/capture.tex
\begin{figure}[t]
\centering
\footnotesize
\setlength{\tabcolsep}{1pt}
\begin{tabular}{cc}
    \includegraphics[height=4.6cm]{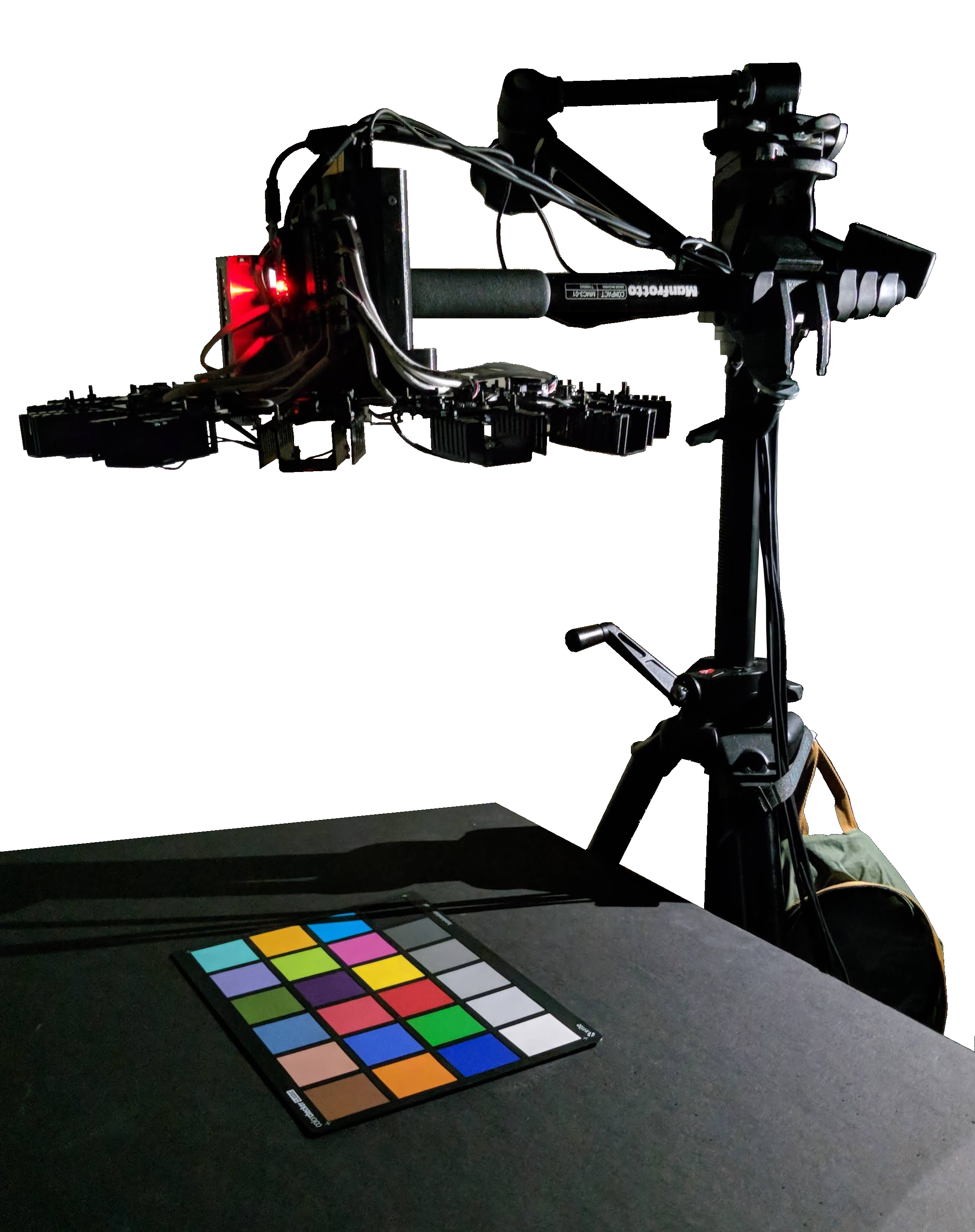} &
    \includegraphics[height=4.6cm]{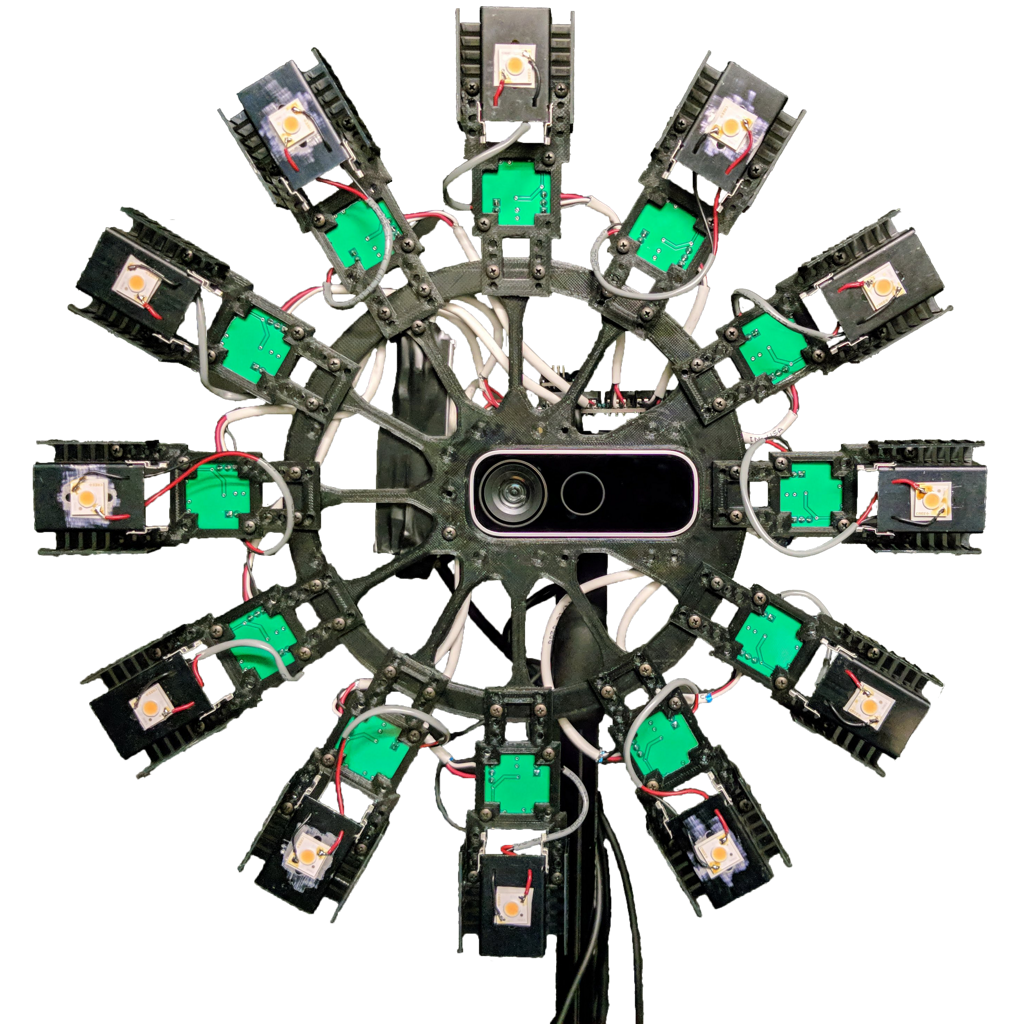} \\
    (a) side view & (b) front view
\end{tabular}
\caption{
Our proposed multi-light capture system shown from the (a) side and (b) front. It consists of 12 LEDs mounted on a 3D printed frame, and surrounding a Kinect Azure RGB-D camera.}
\label{fig:capture-system}
\end{figure}

\section{Capture system and dataset}
\label{sec:capture}

We present our multi-light capture system and show how it is used to capture a dataset of real-world materials. 

\subsection{Capture system overview}
\label{sec:capture-system}
We propose the multi-light capture system illustrated in fig.~\ref{fig:capture-system}, which is composed of a Kinect Azure RGB-D camera surrounded by a 225~mm radius ring of 12 equally-spaced LEDs. All components are mounted rigidly on a custom 3D-printed frame. The LEDs, numbered from 0 (North) and incrementing clockwise, can be switched on/off individually in a round-robin fashion and in sync with the camera (thanks to the Kinect Azure sync line). LED intensity can be adjusted via a 500~Hz PWM signal. The system can capture 4K videos at a frame rate of up to 15~FPS, where each frame corresponds to a different light direction. In all, a full 12-frame capture can be acquired in less than 1 second. 

Our system bears similarities to the one recently proposed by Schmitt~\etal ~\cite{Schmitt2020CVPR}, albeit for different purposes: \cite{Schmitt2020CVPR} focuses on 3D object reconstruction while we use our setup to acquire a dataset of real-world materials. 


\begin{figure*}[t]
\centering
\def\imgw{0.083}
\input{img/fig_tabular_dataset_real/tabular.tex}
\caption{Example sets of images from our dataset of real-world materials. Images lit from lights 0 (North) to 11 (clockwise) are shown from left to right. 
Overall, 80 different samples were captured at various scales, totaling 462 sets of 12 images each. }
\label{fig:dataset-overview}
\end{figure*}
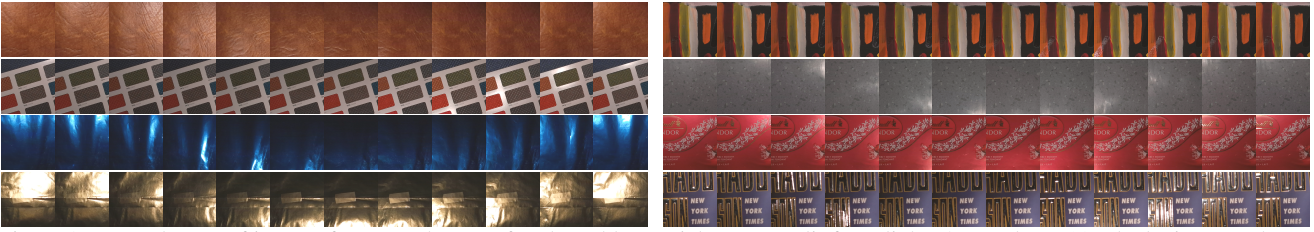

\subsection{A dataset of real-world materials}
\label{sec:dataset}

We leverage our system to capture a novel dataset of real-world materials. First, the system is radiometrically calibrated with an X-Rite ColorChecker and by fitting univariate quadratic functions to each color channel independently. 

A total of 80 real material samples were captured in a dark room. For each material, multiple captures were collected at different distances from the camera (between 250 and 650 mm) to observe both macro- and micro-level details. The dataset is mostly comprised of planar specimens but also includes non-planar objects such as mugs, globes, crumpled paper, etc. As shown in fig.~\ref{fig:dataset-overview}, it contains a rich diversity of materials, including diffuse or specular wrapping papers, fabrics, anisotropic metals, plastics, rugs, ceramic and wood flooring samples, etc. 

Each capture set includes 12 LDR (8 bpp) RGB-D images at 4K pixel resolution. Each set is captured at 50\% and 100\% of maximum light intensity. In total, we captured 462 such image sets (combinations of light intensities, distances to the camera, and material sample). 



%% file: img/fig_tabular_dataset_real/tabular.tex
\begingroup
\def\imgw{0.0410}
\renewcommand{\tabcolsep}{0cm}
\renewcommand*{\arraystretch}{0.0} 
\newcommand{\myimg}[2]{\includegraphics[width=\imgw\linewidth]{img/fig_tabular_dataset_real/img/#1_l#2.png}}
\newcommand{\myimgrow}[1]{\myimg{#1}{03} & \myimg{#1}{02} & \myimg{#1}{01} & \myimg{#1}{00} & \myimg{#1}{11} & \myimg{#1}{10} & \myimg{#1}{09} & \myimg{#1}{08} & \myimg{#1}{07} & \myimg{#1}{06} & \myimg{#1}{05} & \myimg{#1}{04}}

\begin{tabular}{cccccccccccc cccccccccccc}
\myimgrow{038}\hspace{0.6em}\myimgrow{042}\\*[0.1em]
\myimgrow{077}\hspace{0.6em}\myimgrow{099}\\*[0.1em]
\myimgrow{102}\hspace{0.6em}\myimgrow{136}\\*[0.1em]
\myimgrow{030}\hspace{0.6em}\myimgrow{086}\\*[0.1em]
\end{tabular}
\endgroup


%% file: content/method.tex
\section{SVBRDF estimation}

In this section, we describe our approach for SVBRDF estimation. First, we  explain the image formation model used, then describe our differentiable renderer, and finally detail the novel deep learning architecture employed. 

\subsection{Image formation model}
\label{sec:ifm}

Similar to previous work in material estimation~\cite{Valentin_2018,Li_2018,Valentin_2019}, we use the Cook-Torrance microfacet specular shading model~\cite{DisneyBRDF,UnrealBRDF}. Here, lower case symbols refer to scalars while lower case bold symbols refer to $3 \times 1$ column vectors. 


Let $\mathbf{n}_i$, $\mathbf{d}_i$, $r_i$, and $\mathbf{s}_i$ be the surface normal, diffuse albedo, roughness and specular albedo at pixel $i$ in the image. The BRDF model is defined as
\begin{equation}
\rho(\mathbf{n}_i, \mathbf{d}_i, r_i, \mathbf{s}_i) = \frac{\mathbf{d}_i(1-\mathbf{s}_i)}{\pi} 
+ \frac{D(\mathbf{n}_i, r_i) \, F(\mathbf{s}_i) \, G(\mathbf{n}_i, r_i)}{4(\mathbf{n}_i \cdot \mathbf{v}_i)(\mathbf{n}_i \cdot \mathbf{l}_i)} \,,
\end{equation}
where $\mathbf{v}_i$, $\mathbf{l}_i$ are the view and light direction unit vectors. The terms $D$ (GGX/Trowbridge-Reitz normal distribution function~\cite{GGX,TrowbridgeReitz}), $F$ (Schlick function for Fresnel reflection coefficient~\cite{SchlickBRDF}), and $G$ (Schlick-GGX geometric attenuation~\cite{UnrealBRDF}) are defined as

\begin{equation}
\begin{split}
D(\mathbf{n}, r) 	&= \frac{1}{\pi}\left(\frac{r^2}{(\mathbf{n} \cdot \mathbf{h})^2(r^4-1)+1} \right)^2 \,, \\
F(\mathbf{s}) 		&= \mathbf{s} + (1-\mathbf{s})(\mathbf{v}\cdot \mathbf{h})^5  \,, \\
G(\mathbf{n}, r) 	&= \frac{\mathbf{n}\cdot \mathbf{v}}{(\mathbf{n}\cdot \mathbf{v})(1-\frac{r^2}{2}) + \frac{r^2}{2}}\frac{\mathbf{n}\cdot \mathbf{l}}{(\mathbf{n}\cdot \mathbf{l})(1-\frac{r^2}{2}) + \frac{r^2}{2}} \,,
\end{split}
\end{equation}
where $\mathbf{h}_i$ is the half angle unit vector. 

The intensity $\mathbf{b}_i$ of pixel $i$ lit by a point light source of intensity $\mathbf{i}$ and incoming direction $\mathbf{l}$ is formulated as:
\begin{equation}
\mathbf{b}_i = \mathbf{i}  (\mathbf{n}_i \cdot \mathbf{l}) \rho(\mathbf{n}_i, \mathbf{d}_i, r_i, \mathbf{s}_i) \,.
\label{eqn:image-formation}
\end{equation}
%


\subsection{Differentiable renderer}
\label{sec:differentiable-renderer}


We aim to use the reflectance model described in sec.~\ref{sec:ifm} to train our learning-based method. To this end, we create a differentiable renderer, using Pytorch~\cite{Pytorch} to implement eq.~(\ref{eqn:image-formation}). In order to match the real-world capture system from sec.~\ref{sec:capture-system}, the renderer is calibrated as follows. 

The captured images are center-cropped with a size of $512 \times 512$ downsampled to $256 \times 256$. The renderer assumes an orthographic camera---a reasonable assumption given the 28$^{\circ}$ effective field of view of the camera at that resolution.  
A planar surface is placed at a distance corresponding to the mean depth of the patch as measured by the Kinect\footnote{The depth map is used only to obtain the surface distance.} (and thus varies across the dataset) and its dimensions are automatically adjusted to fill the entire field of view of the virtual camera. Virtual point light sources are placed evenly on a circle of 225~mm radius (see sec.~\ref{sec:capture-system}). Renders are performed at a $256 \times 256$ pixel resolution.  

The intensity of each light source is obtained by capturing a calibrated photographer gray card (18\% reflectance). The optimal intensity is found by minimizing the L1 loss between renders of a synthetic graycard material (assumed perfectly Lambertian) and their corresponding real images. 






\subsection{Model architecture}
\label{sec:model-architecture}

\begin{figure}[t]
\centering
\footnotesize
\includegraphics[width=0.75\linewidth]{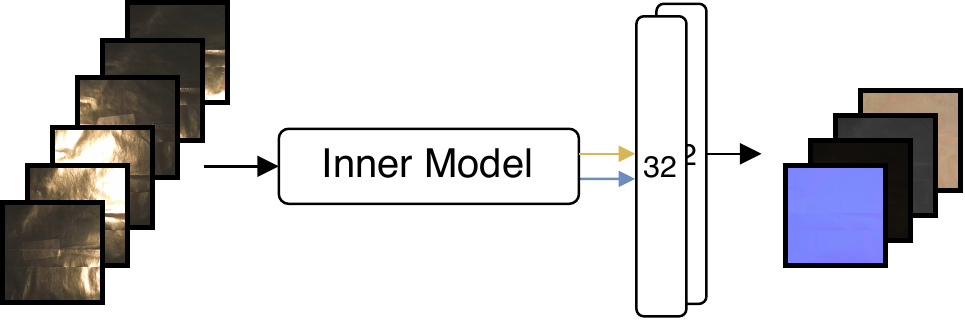}\\
(a) ``Fixed'' model \\
\includegraphics[width=\linewidth]{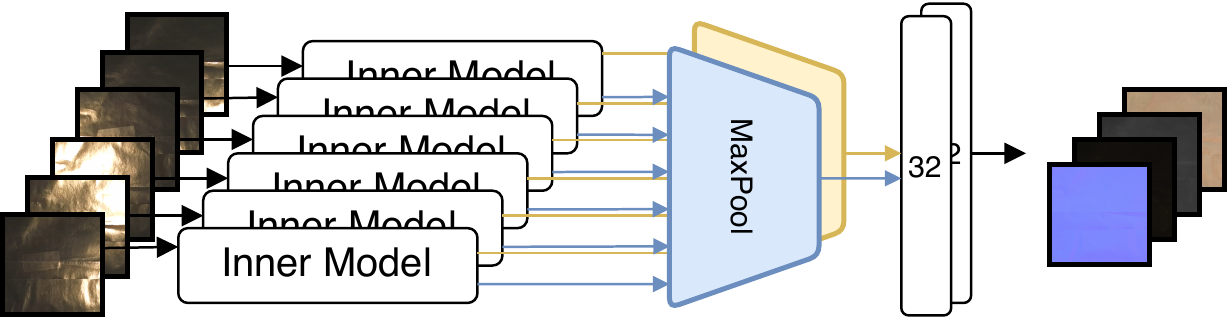} \\ 
(b) ``Dynamic'' model \\
\caption[]{Overview of the two deep learning architectures used in this paper. (a) The ``fixed'' model accepts a fixed number input images (each lit with a different lighting direction), which are concatenated and fed to a single ``inner'' model (fig.~\ref{fig:model_inner}). (b) The ``dynamic'' model accepts a varying number of input images, each image being processed separately by the ``inner'' model (weights are shared across all inner models). Maxpooling is then performed across inner model outputs, separately for each pixel and channel. A fixed model requires much less memory than a dynamic model but can only process a set number of light positions.}
\label{fig:model_regular_vs_maxpool}
\end{figure}

We experiment with two general deep learning architectures for the estimation of SVBRDFs from a varying number of input images (fig.~\ref{fig:model_regular_vs_maxpool}). Each one accepts $K$ input RGB images of a sample at $256 \times 256$ resolution, and outputs the SVBRDF maps as a multi-channel tensor of the same spatial dimensions containing surface normals $\mathbf{N}$, diffuse albedo $\mathbf{D}$, roughness $\mathbf{R}$, and specular albedo $\mathbf{S}$.

The ``fixed'' model (fig.~\ref{fig:model_regular_vs_maxpool}-(a)), accepts a \emph{fixed} number of RGB images, which are concatenated channel-wise into a single tensor, as input and processes this tensor with the ``inner'' model. Intuitively, the ``fixed'' architecture might benefit from the predetermined ordering of input lights.
In contrast, the ``dynamic'' model (fig.~\ref{fig:model_regular_vs_maxpool}-(b)), similar to \cite{Valentin_2019}, is fed with a \emph{varying} number of input images processed separately by several instances of an ``inner'' model. Max-pooling is then performed channel-wise on the resulting feature maps. The outputs of both models are processed by a Styled\-Conv\-Block (fig.~\ref{fig:style_block}), to produce the SVBRDF maps. 

\begin{figure}[t]
\centering
    \includegraphics[width=\linewidth]{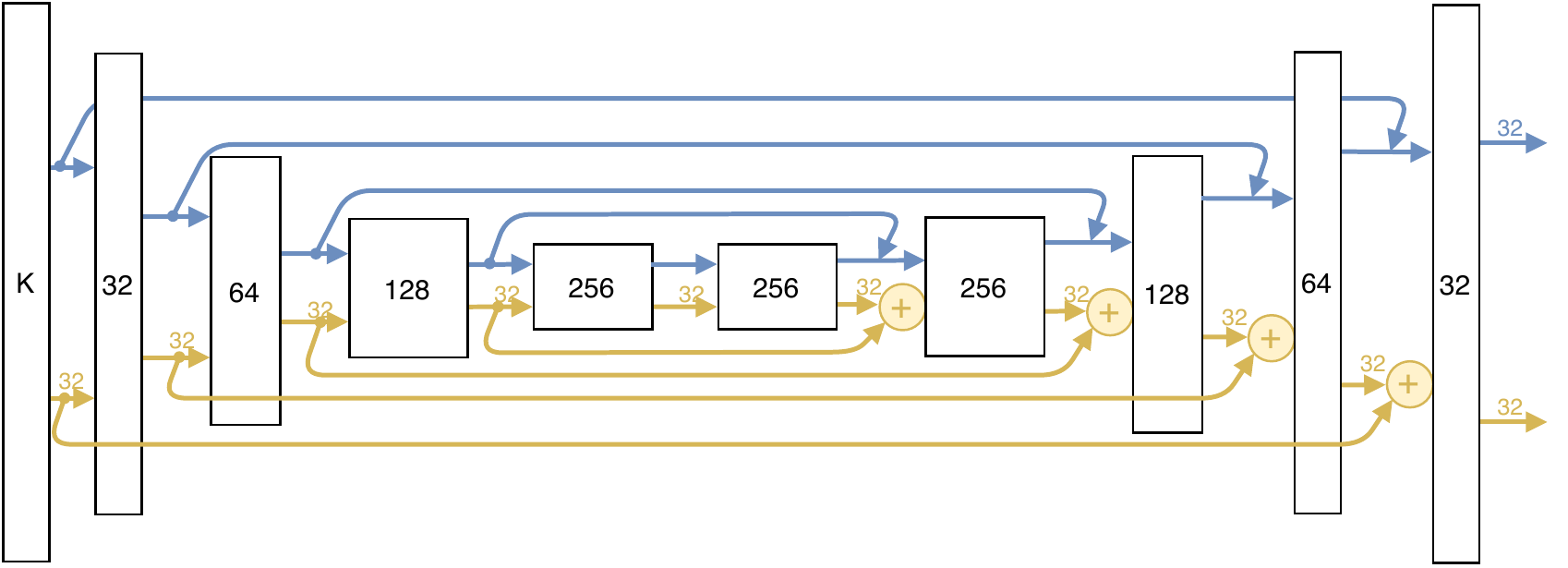}
\caption{Inner model overview. Links for style features (yellow) are additions and links for convolutional features (blue) are concatenations. The numbers in the blocks represent output convolutional channels. The encoding blocks begin with a 2x downsampling operation. The decoding blocks begin with a 2x upsampling operation. The internal structure of a block is described in fig.~\ref{fig:style_block}.}
\label{fig:model_inner}
\end{figure}

\begin{figure}[t]
\centering
    \includegraphics[width=\linewidth]{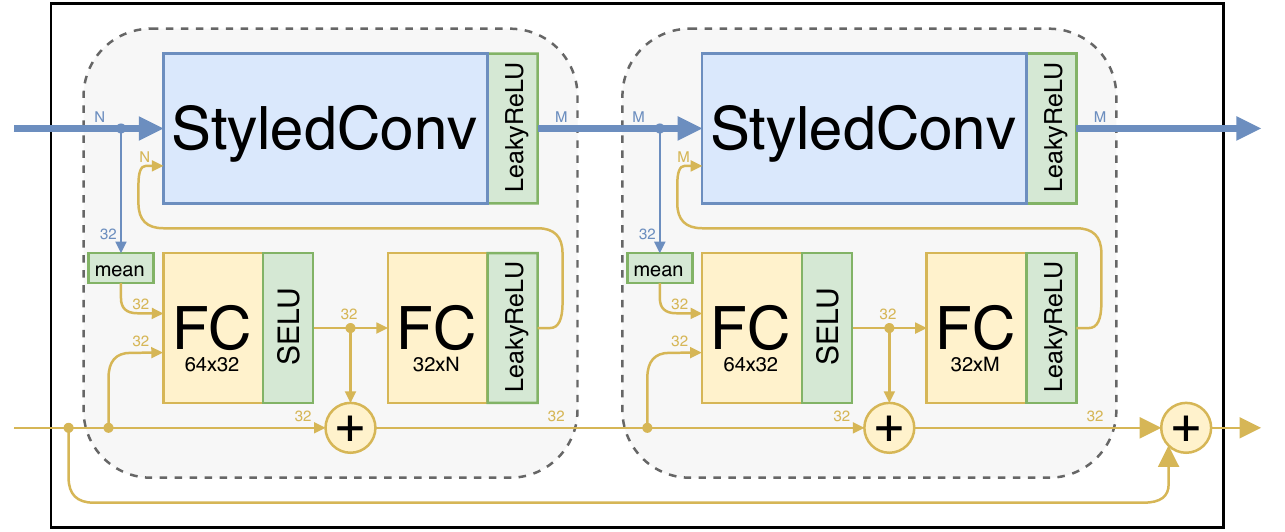}
\caption{This figure presents details of a block of the inner model shown in fig. \ref{fig:model_inner}. The style track input (yellow) is concatenated to the mean of the first 32 convolutional channels (blue) and fed to the first FC layer. Using both the mean and the style input allows some information to transfer back to the style track. As in \cite{Karras2019stylegan2}, StyledConv are 2D convolutions with bias where weights are updated on-the-fly according to $\mathbf{w}'= \mathbf{w}^T \sigma$. This $\sigma$ (style) input to the convolution comes from the second FC layer.}
\label{fig:style_block}
\end{figure}

Both proposed networks rely on the same ``inner model'', illustrated in fig.~\ref{fig:model_inner}, which is based on the U-Net convolutional architecture~\cite{ronneberger2015u} (blue arrows in fig.~\ref{fig:model_inner}). In addition, we follow \cite{Valentin_2018,karras2019style} and complement the main convolutional path with a parallel ``style'' track. Whereas \cite{Valentin_2018,Valentin_2019} use this track to inject material information (style) with adaptive instance normalization (AdaIN)~\cite{AdaIN}, we have observed, as in \cite{karras2019style}, that this tends to create water-like droplet artifacts. We thus follow \cite{karras2019style} and replace the normalization layers with weaker regularization through the use of the ``StyledConv'' layers, the details of which are shown in fig.~\ref{fig:style_block}. 
As an added benefit, this approach yields a smaller network (6.4M parameters, 10x fewer than \cite{Valentin_2019}), which in turn enables training with larger batch sizes. 




\section{Training procedure}
\label{sec:training}

As with previous work, the network is first trained on synthetic data. Then, two options are investigated to adapt the network to real-data: 1) generic finetuning; and 2) material-specific optimization. 

\subsection{Training on synthetic materials}
\label{sec:training-synthetic}

\begin{figure}[t]
\centering
    \includegraphics[width=\linewidth]{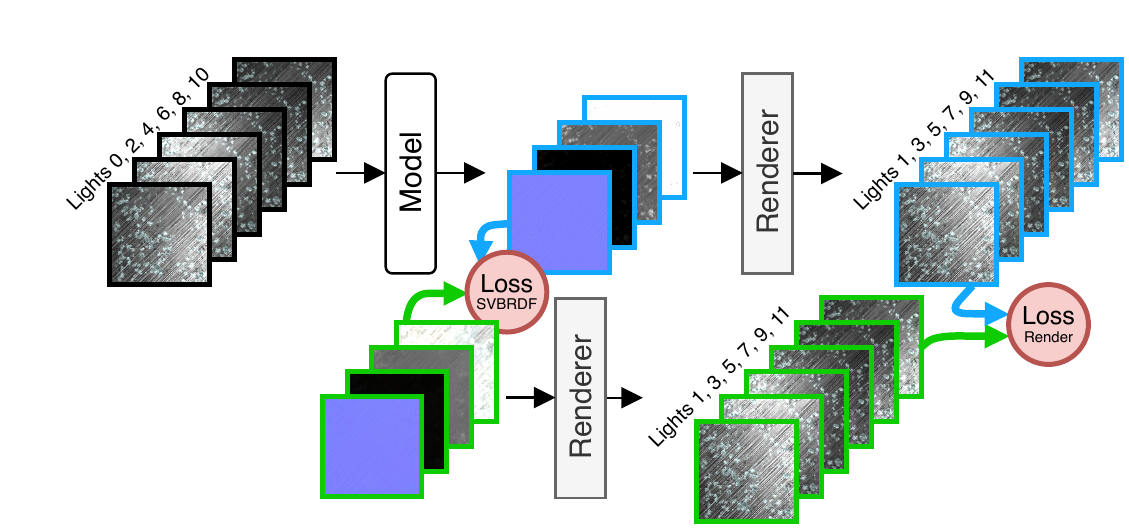}
\caption{Overview of the training procedure on synthetic data. Renders (black), generated from ground truth SVBRDF (green), are fed to a model. The output of the model (blue) is compared to the ground truth SVBRDF. Both predicted and ground truth SVBRDF maps are then used with novel light positions to generate renders which are compared to compute the render loss. }
\label{fig:training_synthetic_vs_regular}
\end{figure}

We borrow the training procedure and synthetic dataset of Deschaintre~\etal~\cite{Valentin_2018,Valentin_2019}. For completeness, we briefly explain the main steps here, but refer the interested reader to the aforementioned references for more details. 

A large dataset of synthetically-generated SVBRDFs is used to render synthetic training images by lighting them using our differentiable renderer (sec.~\ref{sec:differentiable-renderer}). For the purpose of data augmentation, material mixing is performed on the fly so SVBRDF maps of two random materials are combined to become a new valid SVBRDF map. In addition, we randomly flip, mirror and crop a $256 \times 256$ image from the original $512 \times 512$ material.

The overall training procedure is illustrated in fig.~\ref{fig:training_synthetic_vs_regular}. Ground truth SVBRDF maps are used to render input images (left, black outline). These synthetic renders are then fed to the network to generate a predicted SVBRDF map (middle, blue outline). As proposed by \cite{Li:2017:MSA}, the overall loss function between a predicted SVBRDF map $\tilde{\mathbf{B}}$ and the ground truth $\mathbf{B}$ is a sum of two losses
\begin{equation}
\mathcal{L}(\tilde{\mathbf{B}}, \mathbf{B}) = \mathcal{L}_\mathrm{brdf}(\tilde{\mathbf{B}}, \mathbf{B}) + \mathcal{L}_\mathrm{render}(\tilde{\mathbf{B}}, \mathbf{B}) \,.
\label{eqn:loss}
\end{equation}
First, each of the predicted SVBRDF maps are individually compared to the ground truth (middle, green outline in fig.~\ref{fig:training_synthetic_vs_regular}) using the $L_1$ loss in eq. \ref{eqn:svbrdf-loss}.
These weights were found empirically and do not have a strong influence the final results on real materials since this loss is active only in the pre-training phase on synthetic materials. 
\begin{equation}
\begin{split}
\mathcal{L}_\mathrm{brdf}(\tilde{\mathbf{B}}, \mathbf{B}) & = 
	10 \mathcal{L}_1(\tilde{\mathbf{N}}, \mathbf{N}) 
	+ 3 \mathcal{L}_1(\tilde{\mathbf{D}}, \mathbf{D}) \\
	& + \mathcal{L}_1(\tilde{\mathbf{R}}, \mathbf{R})
	+ 2 \mathcal{L}_1(\tilde{\mathbf{S}}, \mathbf{S}) \,.
\end{split} 
\label{eqn:svbrdf-loss}
\end{equation}
Second, the differentiable renderer $\mathcal{R}(\cdot)$ (see sec.~\ref{sec:differentiable-renderer}), which implements eq.~(\ref{eqn:image-formation}), is used to generate a set of $K$ renders with light directions $\mathbf{l}_i, i=1 \ldots K$ (right in fig.~\ref{fig:training_synthetic_vs_regular}, blue outline), which are compared to renders of the ground truth SVBRDF maps (right, green outline) with the same light directions. They are compared using the $L_1$ loss after being tonemapped to logarithmic space to flatten the dynamic range and mitigate the influence of specular peak errors as in \cite{Valentin_2018,Valentin_2019} with $tm(x) = \frac{\log(x+0.01) - \log(0.01)}{\log(1.01)-\log(0.01)}$:
\begin{equation}
\mathcal{L}_\mathrm{render}(\tilde{\mathbf{B}}, \mathbf{B}) = 
	\sum_{i=1}^K \mathcal{L}_1(tm(\mathcal{R}(\tilde{\mathbf{B}}, \mathbf{l}_i)), tm(\mathcal{R}(\mathbf{B}, \mathbf{l}_i))) \,.
\label{eqn:render-loss}
\end{equation}

Different from \cite{Valentin_2018} in which light positions are sampled randomly, we instead mimic the geometric configuration of our capture device (sec.~\ref{sec:capture-system}). The distance to the specimens is sampled using a uniform distribution in the range $[250, 650]$~mm, which represents the depth of captures in the real dataset. Virtual point light sources are evenly placed on a circle of 225~mm radius, and noise is added to the input renders light positions using random offsets in both $x$ and $y$ directions sampled from a uniform distribution in the range $[-20, 20]$~mm. As illustrated in fig.~\ref{fig:training_synthetic_vs_regular}, even-numbered lights are used as input, and odd-numbered lights for computing the render loss $\mathcal{L}_\mathrm{render}$. The Adam optimizer \cite{kingma2014adam} is used with default parameters and a batch size of 4.

\subsection{Generic finetuning}
\label{sec:training-finetuning}

A natural approach for bridging the gap between the synthetic and real domain is to finetune the network using a small subset of real training data. Since acquiring ground truth SVBRDF maps requires complex capture apparatus, we instead rely on our dataset and finetune the network exclusively using the render loss $\mathcal{L}_\mathrm{render}$ from eq.~(\ref{eqn:render-loss}). Again, even- and odd-numbered lights are respectively used as input and for computing the loss. 

Models are trained on a 50/50 split of the real material dataset, for a total of 100 epochs, using the Adam optimizer and a batch size of 4. The train/test split is done manually to ensure no similar samples appear in the train and test set. 

\subsection{Material-specific optimization}
\label{sec:training-optimization}

An alternative approach---which can also be applied in conjunction with finetuning---is to optimize the network weights specifically for a single material. Similar to recent work reported in \cite{Deschaintre2020}, this is essentially equivalent to ``overfitting'' to the input images of a single material. 
We allow the model to train, again with the Adam optimizer, on a single material for 100 additional iterations using the render loss $\mathcal{L}_\mathrm{render}$ from eq.~(\ref{eqn:render-loss}) only. 
Here, the even-numbered lights are used both for input and for computing the loss. Odd-numbered lights will be used to report performance. 


%% file: content/experiments.tex
\section{Experiments}

In this section, we evaluate and compare the networks of sec.~\ref{sec:training} on synthetic and real data. 
Unless otherwise noted, all trained networks take as input 6 images, corresponding to the even-numbered light sources, from a test material (unseen during training). Performance is then reported on odd-numbered light sources. 


\subsection{Synthetic data} 
\label{sec:synthetic-experiments}

\begin{figure}[t]
\centering
  \includegraphics[width=0.9\linewidth]{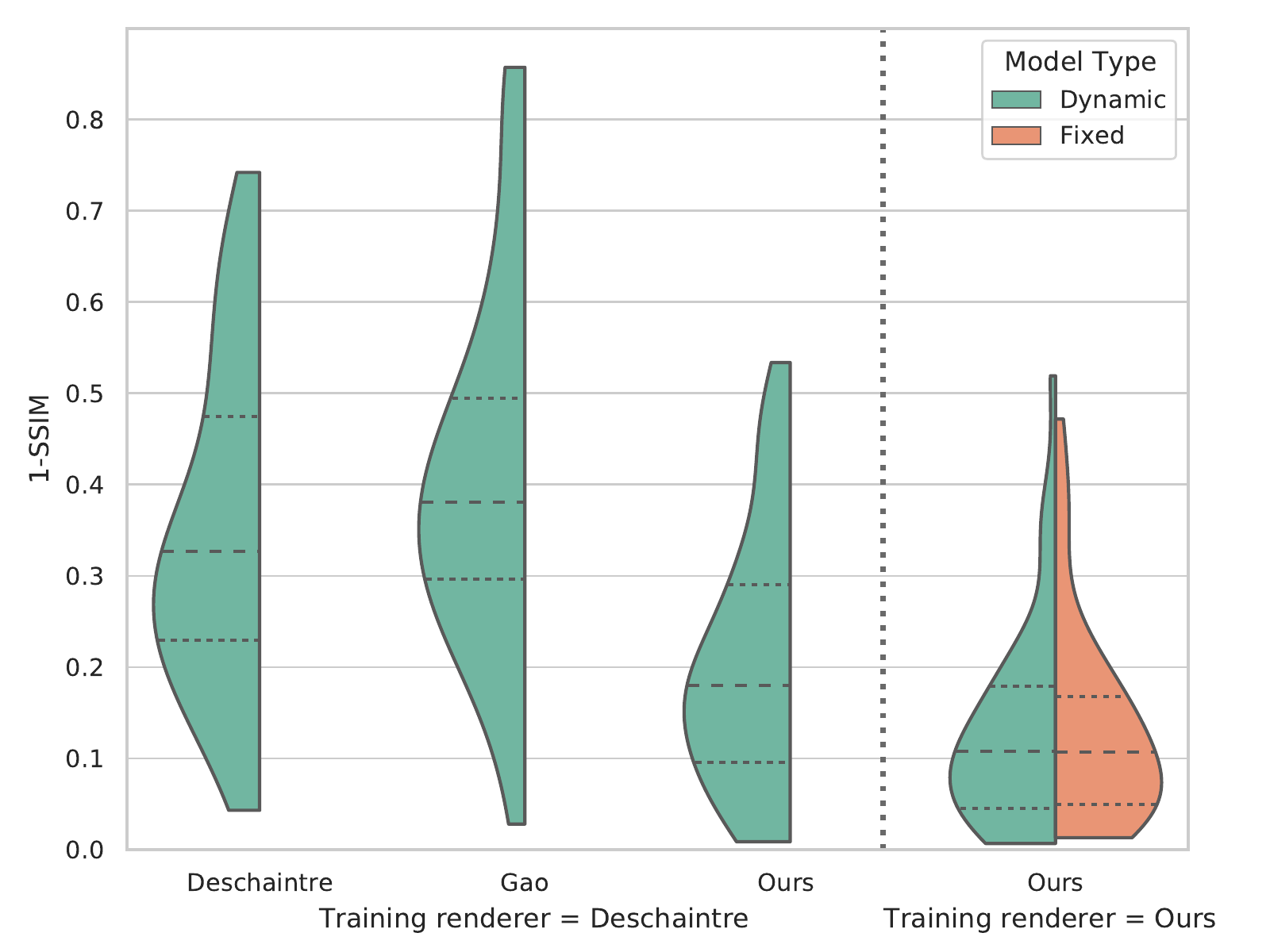}
\caption{Quantitative comparison between our networks and previous work (Deschaintre~\etal~\cite{Valentin_2019} and Gao~\etal~\cite{Gao_2019}) using 1-SSIM (lower is better) on the \emph{synthetic} test set. The models on the left of the vertical dotted line are trained with the renderer from Deschaintre~\etal~\cite{Valentin_2019}. On the right are models trained with our renderer and a configuration matching our capture system  (sec.~\ref{sec:differentiable-renderer}). All models take 6 input images and are evaluated using our renderer and configuration. The loss is computed on novel renders. 
}
\label{fig:violin_synthetic}
\end{figure}

This first experiment compares networks trained on the synthetic dataset (see sec.~\ref{sec:training-synthetic}), the results of which are shown in fig.~\ref{fig:violin_synthetic}. Here, we compare two options for rendering the images used as input and for computing the loss. First, we mimic the configuration proposed in Deschaintre~\etal~\cite{Valentin_2019} (also used by Gao~\etal~\cite{Gao_2019}), in which the number of input images, view direction, light directions and intensity are sampled randomly. The results are shown on the left side of fig.~\ref{fig:violin_synthetic}. Second, renders are produced using our light configuration (sec.~\ref{sec:training-synthetic}). These results are shown on the right side of fig.~\ref{fig:violin_synthetic}. 



Fig.~\ref{fig:violin_synthetic} shows that, when trained with the Deschaintre renderer, our architecture, despite requiring much fewer parameters, achieves improved performance according to the 1-SSIM metric. In our experiments, the method from Gao~\etal~\cite{Gao_2019} resulted in worse performance despite a longer process of optimization and refinement. Furthermore, our model performs even better when using both our renderer and light configuration. Interestingly, the ``dynamic'' and ``fixed'' architectures (c.f. sec.~\ref{sec:model-architecture}) perform quite similarly. Indeed, the flexibility of the ``dynamic'' architecture (since it accepts a variable number of inputs) does not seem to hamper its performance. 
Please consult the supplementary for other metrics (L1, perceptual~\cite{zhang2018unreasonable}) for this experiment.



\begin{figure}[t]
\centering
    \includegraphics[width=0.9\linewidth]{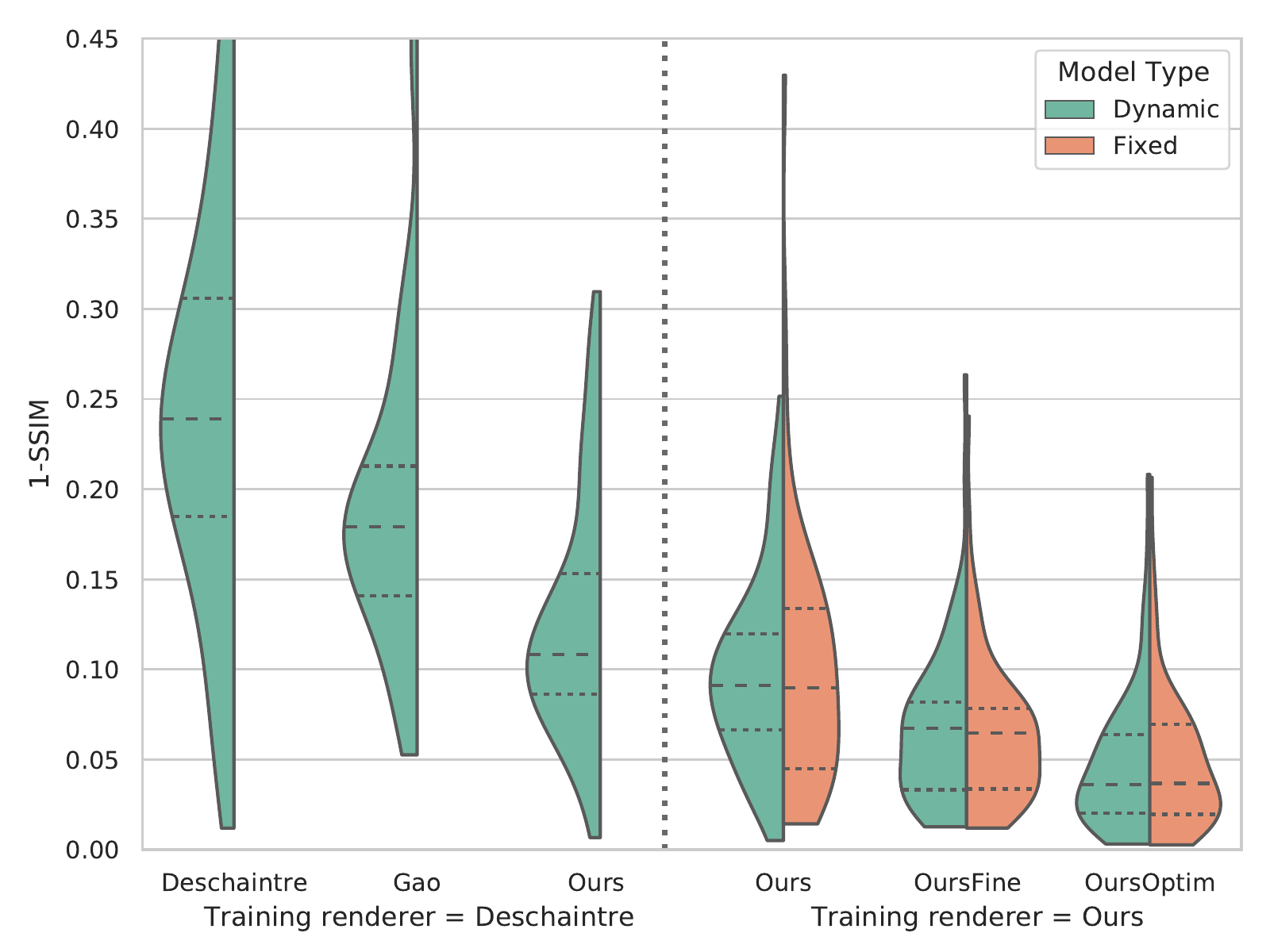}
\caption{Quantitative comparison between our networks and previous work (Deschaintre~\etal~\cite{Valentin_2019} and Gao~\etal~\cite{Gao_2019}) using 1-SSIM (lower is better) on the \emph{real} test set. 
Six images are fed to the model which predicts SVBRDF parameters. 
Deschaintre's renderer (left) use random unseen light position at test time, while ours (right) use the even-indexed light sources on the capture rim for training and odd-indexed lights for test.
All methods are trained on synthetic images. Only ``OursFine'' and ``OursOptim'' are finetuned on real images. See sec.~\ref{sec:model-architecture} and~\ref{sec:training} for more details.
}
\label{fig:violin_real}
\end{figure}

\def\imgw{0.138}

\begin{figure}[t]
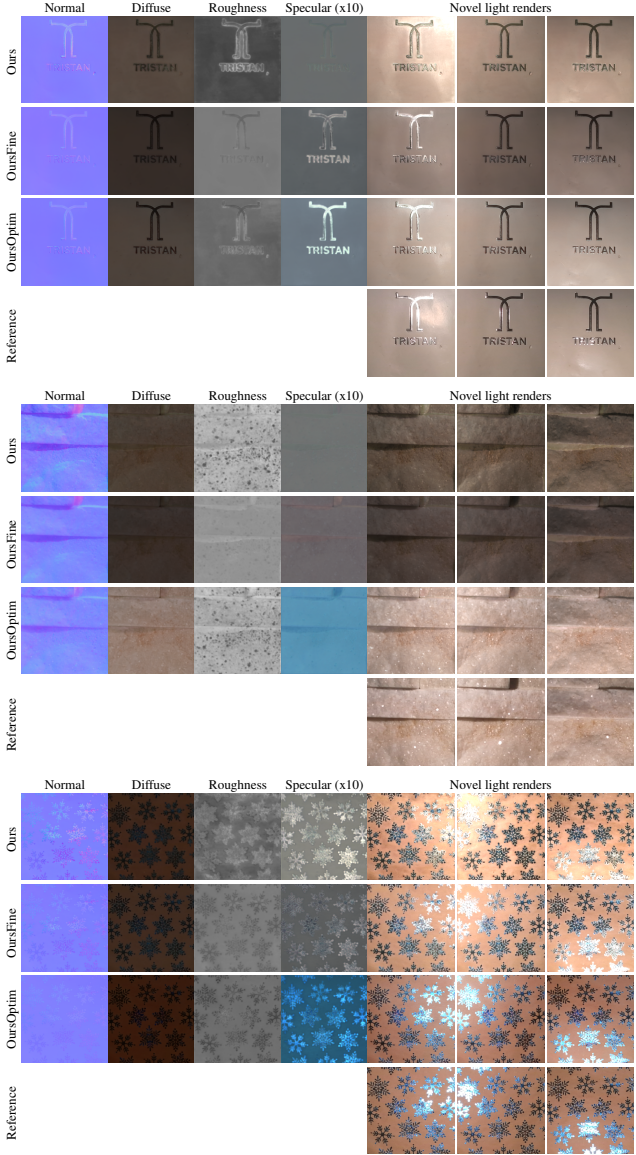

  \def\imgw{0.138} 
\include{img/fig_tabular_max_material_9/tabular}
\include{img/fig_tabular_max_material_13/tabular}
\include{img/fig_tabular_max_material_54/tabular}
  \caption{Qualitative comparison of training on synthetic data (``Ours''), generic finetuning (``OursFine'') and per-material optimization (``OursOptim''). Here, the ``dynamic'' model is used. The last row shows the ground truth (SVBRDF maps unavailable). }
\label{fig:fine_optim}
\end{figure}

\begin{figure}[t]
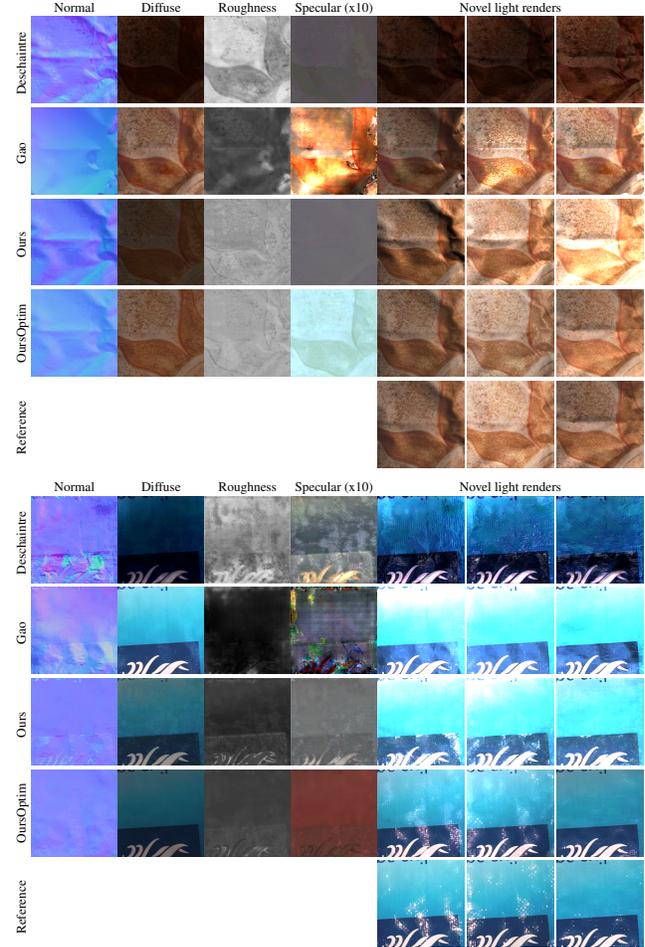

  \def\imgw{0.138} 
\include{img/fig_tabular_gao_material_33/tabular}
\include{img/fig_tabular_gao_material_6/tabular}
  \caption{Qualitative comparison between previous work (Deschaintre~\etal~\cite{Valentin_2019}, Gao~\etal~\cite{Gao_2019}) and our model trained on synthetic data only (``Ours''), and after the per-material optimization (``OursOptim''). Here, the ``dynamic'' model is used. The last row shows the ground truth (SVBRDF maps unavailable).}
\label{fig:previous_work_compare_synth}
\end{figure}

\subsection{Applying the synthetic model on real images}

We now evaluate the models trained on synthetic data and tested on our real dataset, which follows what is proposed in several previous works~\cite{Valentin_2018,Li_2018,Valentin_2019, Gao_2019}. Quantitative and qualitative results are reported in figs \ref{fig:violin_real} and \ref{fig:previous_work_compare_synth} respectively. As in sec.~\ref{sec:synthetic-experiments} and fig.~\ref{fig:violin_synthetic}, training with the Deschaintre~\cite{Valentin_2018} and our renderer are both evaluated. As before, our architecture outperforms previous approaches in both cases. 


\subsection{Real data adaptation} 

We now compare networks trained on synthetic data only with those finetuned on real data. The results are shown quantitatively in fig.~\ref{fig:violin_real}, and qualitatively in figs.~\ref{fig:fine_optim}. 
The ``OursFine'' column in fig.~\ref{fig:violin_real} shows the results of finetuning the networks on the real training set (c.f. sec.~\ref{sec:training-finetuning}). Compared to the version trained on synthetic data only (``Ours'' label), the finetuning reduces the median 1-SSIM measure by 0.02 from 0.09 to 0.07. While this difference may not appear as significant, it actually achieves a visible improvement on the results, as can be seen when comparing the dynamic and finetuned rows of fig.~\ref{fig:fine_optim}. This shows the significance of the gap between both domains, and invalidates the common hypothesis of training on synthetic data only. 

The ``OursOptim'' column in fig.~\ref{fig:violin_real} shows the results obtained by performing the per-material optimization scheme from sec.~\ref{sec:training-optimization}. This is done individually for each test material, and results are aggregated over all materials. Compared to the finetuned results, doing so ends up in a significant improvement in the (1-SSIM) metric, where the median drops to approximately 0.035. This also translates in much improved visual quality, as shown in fig.~\ref{fig:fine_optim}.  

Fig.~\ref{fig:previous_work_compare_synth} shows qualitative examples comparing our approach (the ``dynamic'' network with finetuning and per-material optimization) with the approaches of Deschaintre~\etal~\cite{Valentin_2019} and Gao~\etal~\cite{Gao_2019}. It can be observed that the normal maps estimated by \cite{Valentin_2019} display more high frequencies, but unfortunately many of these details are wrong: surface normals discontinuities explain changes that should be in other maps. Our ``dynamic'' network qualitatively outperforms the others. There is still a significant difference between the novel light renders and the captured ground truth, justifying the need for adjusting the networks to real data. 

While it achieves the best results, the main downside of the per-material optimization is the lack of efficiency. Indeed, 30 seconds are required to perform the 100 training iterations needed. Consequently, time-critical applications would have to resort to the finetuning approach. 



\begin{figure}[t]
\centering
  \includegraphics[width=0.9\linewidth]{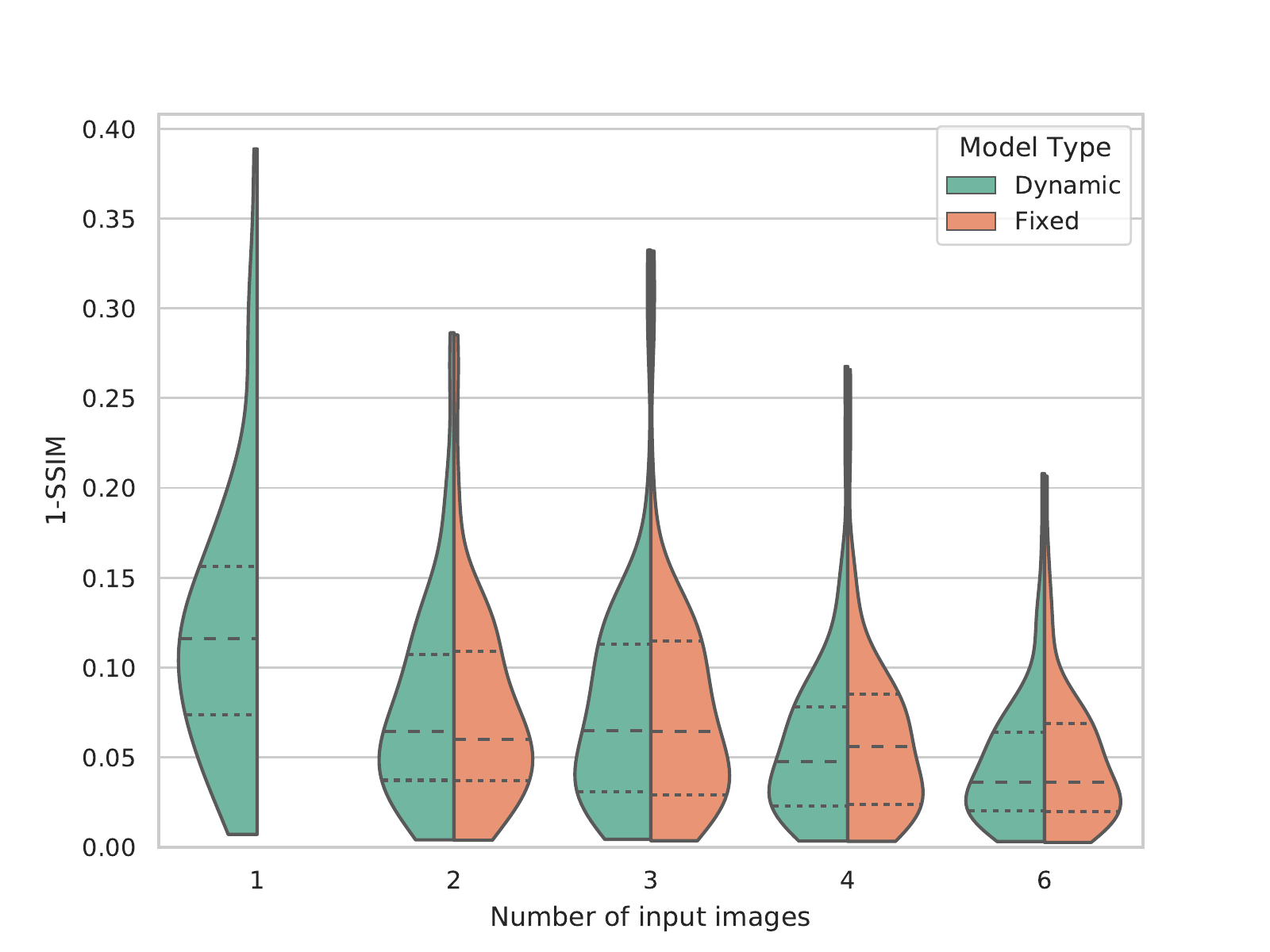}
\caption{Effect of varying the number of input images for training. Here, the per-material optimization results are shown for the ``dynamic'' and ``fixed'' architectures. When the number is 1, both architectures are the same so only one result is shown.}
\label{fig:violin_x_inputs}
\end{figure}

\subsection{Ablations}

\paragraph{Varying number of input light directions}

As \cite{xu2018deep, Valentin_2019} have shown, only a small number of images are required to predict a satisfying SVBRDF. All the experiments presented before were trained and evaluated using six input images. 

We show the results obtained by varying the number of input images from 1 to 6 in fig.~\ref{fig:violin_x_inputs} (all 6 odd-numbered images are still kept for evaluation). As expected, increasing the number of input images helps in reducing the error. The large gap observed when going from one to two images suggests that using a single image as input is still underconstrained and that the network cannot learn sufficiently powerful priors to yield robust results. Qualitatively, we observe that two inputs are generally enough for diffuse materials. For specular or more complex materials, four inputs are good while six are slightly better (see supplementary).

Models with varying number of lights use indices (related to images and their light positions) in the following sequence: $[0, 6, 8, 2, 4, 10]$ (e.g. the models using one input use index $[0]$, and the models using four inputs use indices $[0, 6, 8, 2]$).
Even if some configurations are not symmetrical and unevenly split, this allows a fair comparison between models with a different number of inputs.

\begin{figure}[t]
\centering
  \includegraphics[width=0.9\linewidth]{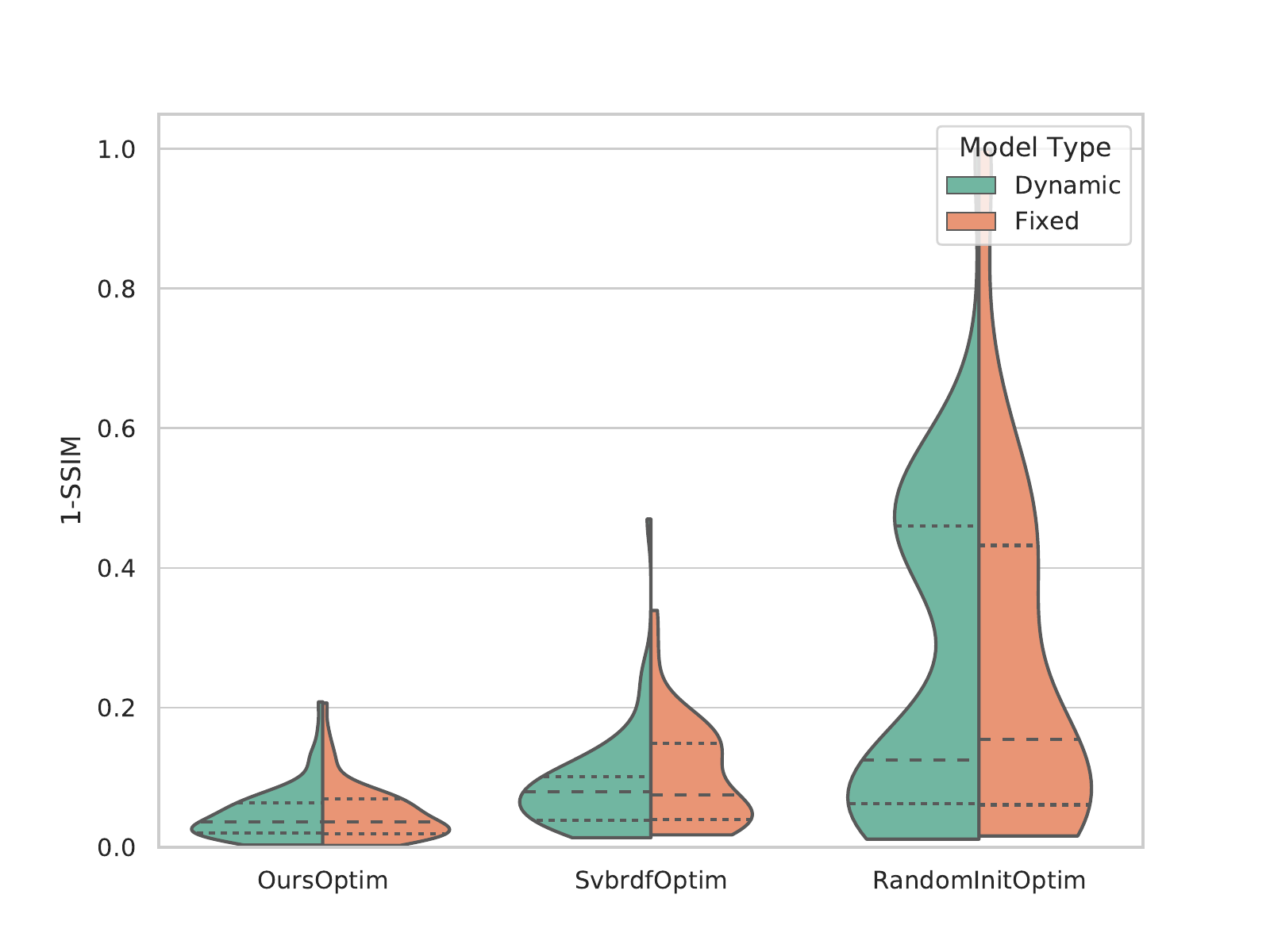}
\caption{Comparison of different approaches for the per-material optimization. All methods minimize $\mathcal{L}_\mathrm{render}$ (eq.~(\ref{eqn:render-loss})) on the input images. ``OursOptim'' overfits the weights of a model trained on the synthetic dataset to the input images. ``RandomInit'' performs the same process but on a randomly initialized and untrained model. ``SvbrdfRefined'' optimizes the SVBRDF maps directly.}
\label{fig:refinement_methods}
\end{figure}

\paragraph{Material-specific optimization methods}

We compare the proposed material-specific optimization, which affects the network weights using the render loss $\mathcal{L}_\mathrm{render}$ (eq.~(\ref{eqn:render-loss})) on the input images, with optimizing the SVBRDF maps directly (without the network). Here, SGD on $\mathcal{L}_\mathrm{render}$ is run for 2000 iterations (with a patience parameter of 100, Adam did not achieve successful results). Results are reported in fig.~\ref{fig:refinement_methods}, and show that the network acts as a strong regularizer which prevents overfitting. Note that, as opposed to \cite{Gao_2019}, there was no need to train a separate autoencoder. Fig.~\ref{fig:refinement_methods} also compares with using an untrained network, which results in much worse performance, thereby validating the need for using trained weights.

%% file: img/fig_tabular_max_material_9/tabular.tex
\tiny
\begin{tabularx}{\linewidth}{@{}X@{} @{}c@{} @{}c@{} @{}c@{} @{}c@{} @{}c@{} }
& Normal & Diffuse & Roughness & Specular (x10) & Novel light renders \\
\rotatebox{90}{\hspace{3.7mm}Ours} & 
\includegraphics[width=\imgw\linewidth]{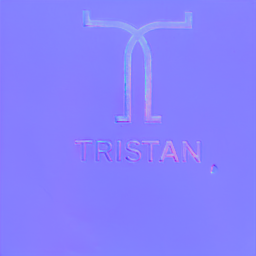} & 
\includegraphics[width=\imgw\linewidth]{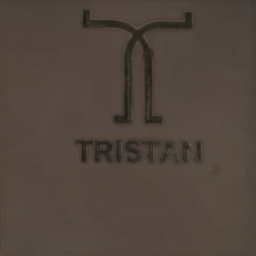} & 
\includegraphics[width=\imgw\linewidth]{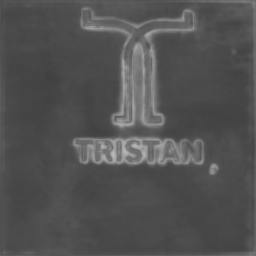} & 
\includegraphics[width=\imgw\linewidth]{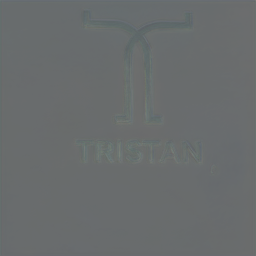} & 
\includegraphics[width=\imgw\linewidth]{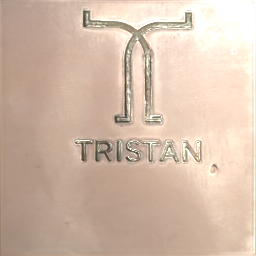}
\includegraphics[width=\imgw\linewidth]{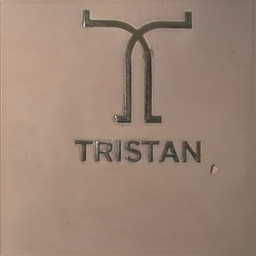}
\includegraphics[width=\imgw\linewidth]{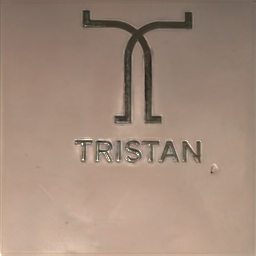}\\
\rotatebox{90}{\hspace{1.8mm}OursFine} & 
\includegraphics[width=\imgw\linewidth]{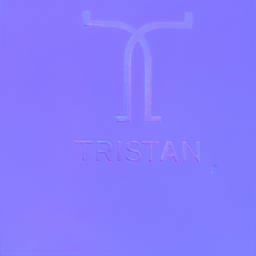} & 
\includegraphics[width=\imgw\linewidth]{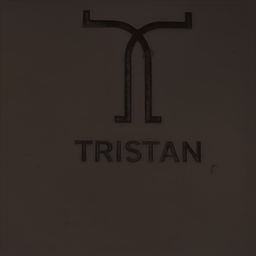} & 
\includegraphics[width=\imgw\linewidth]{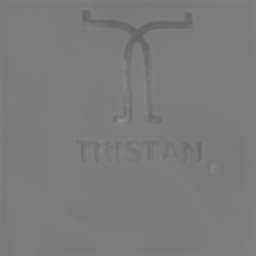} & 
\includegraphics[width=\imgw\linewidth]{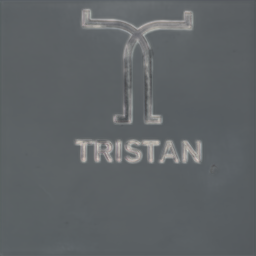} & 
\includegraphics[width=\imgw\linewidth]{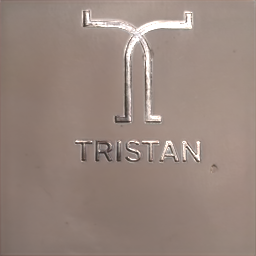}
\includegraphics[width=\imgw\linewidth]{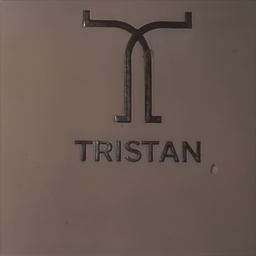}
\includegraphics[width=\imgw\linewidth]{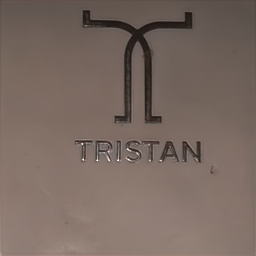}\\
\rotatebox{90}{\hspace{1.8mm}OursOptim} & 
\includegraphics[width=\imgw\linewidth]{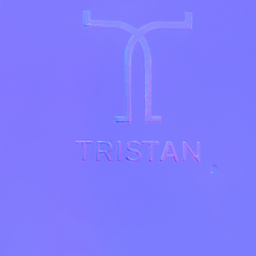} & 
\includegraphics[width=\imgw\linewidth]{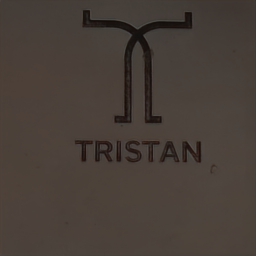} & 
\includegraphics[width=\imgw\linewidth]{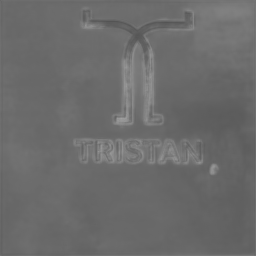} & 
\includegraphics[width=\imgw\linewidth]{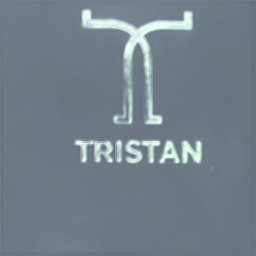} & 
\includegraphics[width=\imgw\linewidth]{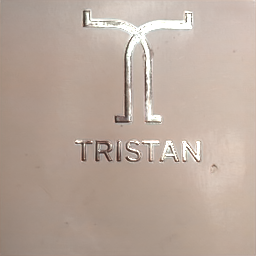}
\includegraphics[width=\imgw\linewidth]{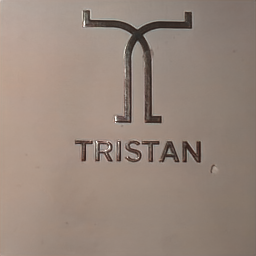}
\includegraphics[width=\imgw\linewidth]{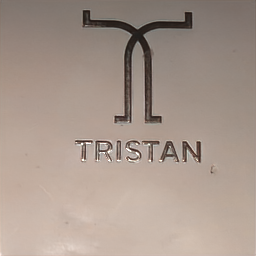}\\
\rotatebox{90}{\hspace{1.8mm}Reference} & 
 & 
 & 
 & 
 & 
\includegraphics[width=\imgw\linewidth]{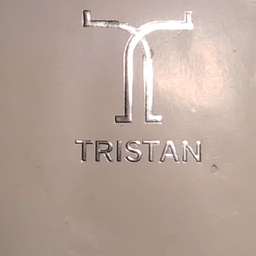}
\includegraphics[width=\imgw\linewidth]{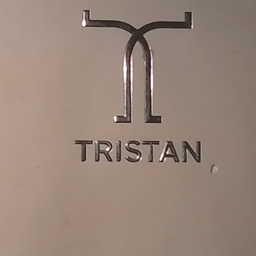}
\includegraphics[width=\imgw\linewidth]{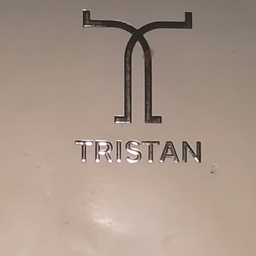}\\

\end{tabularx}

%% file: img/fig_tabular_max_material_13/tabular.tex
\tiny
\begin{tabularx}{\linewidth}{@{}X@{} @{}c@{} @{}c@{} @{}c@{} @{}c@{} @{}c@{} }
& Normal & Diffuse & Roughness & Specular (x10) & Novel light renders \\
\rotatebox{90}{\hspace{3.7mm}Ours} & 
\includegraphics[width=\imgw\linewidth]{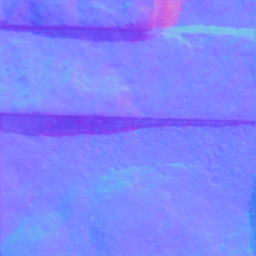} & 
\includegraphics[width=\imgw\linewidth]{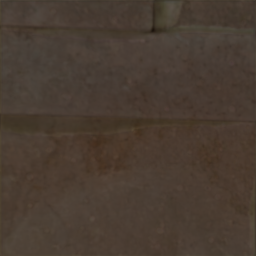} & 
\includegraphics[width=\imgw\linewidth]{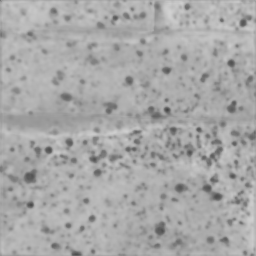} & 
\includegraphics[width=\imgw\linewidth]{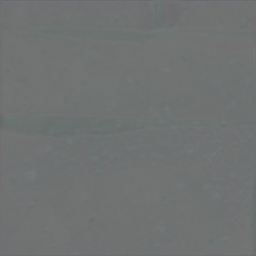} & 
\includegraphics[width=\imgw\linewidth]{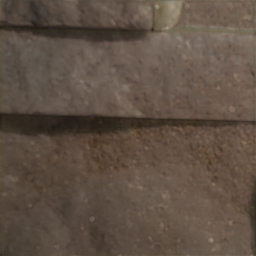}
\includegraphics[width=\imgw\linewidth]{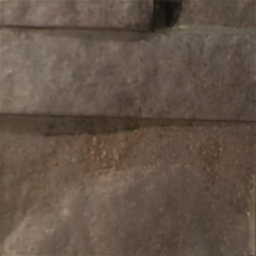}
\includegraphics[width=\imgw\linewidth]{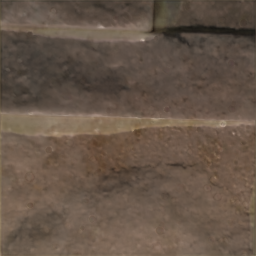}\\
\rotatebox{90}{\hspace{1.8mm}OursFine} & 
\includegraphics[width=\imgw\linewidth]{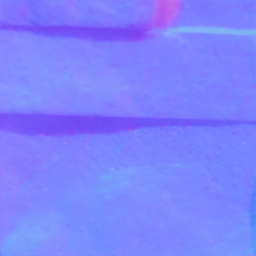} & 
\includegraphics[width=\imgw\linewidth]{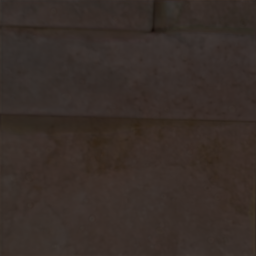} & 
\includegraphics[width=\imgw\linewidth]{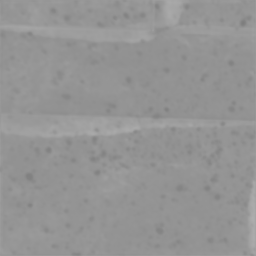} & 
\includegraphics[width=\imgw\linewidth]{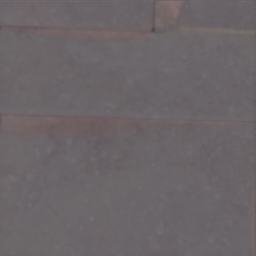} & 
\includegraphics[width=\imgw\linewidth]{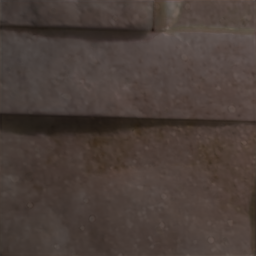}
\includegraphics[width=\imgw\linewidth]{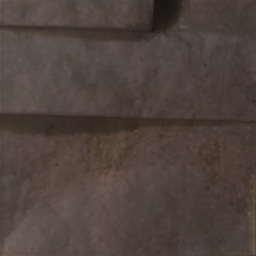}
\includegraphics[width=\imgw\linewidth]{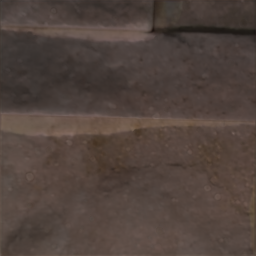}\\
\rotatebox{90}{\hspace{1.8mm}OursOptim} & 
\includegraphics[width=\imgw\linewidth]{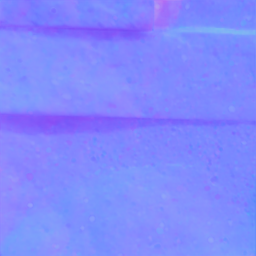} & 
\includegraphics[width=\imgw\linewidth]{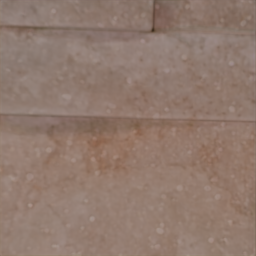} & 
\includegraphics[width=\imgw\linewidth]{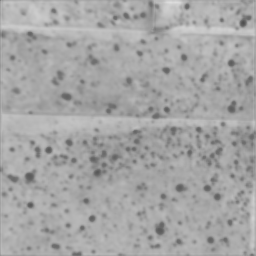} & 
\includegraphics[width=\imgw\linewidth]{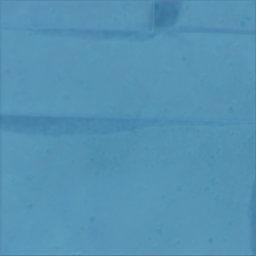} & 
\includegraphics[width=\imgw\linewidth]{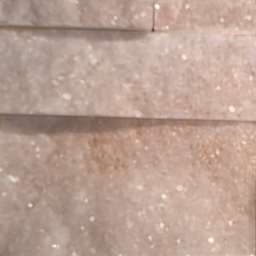}
\includegraphics[width=\imgw\linewidth]{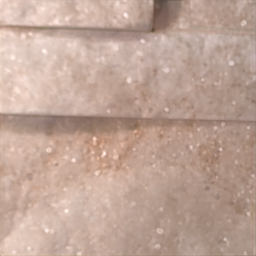}
\includegraphics[width=\imgw\linewidth]{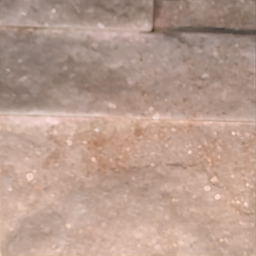}\\
\rotatebox{90}{\hspace{1.8mm}Reference} & 
 & 
 & 
 & 
 & 
\includegraphics[width=\imgw\linewidth]{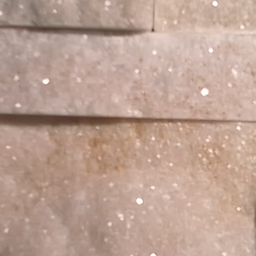}
\includegraphics[width=\imgw\linewidth]{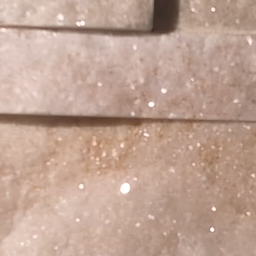}
\includegraphics[width=\imgw\linewidth]{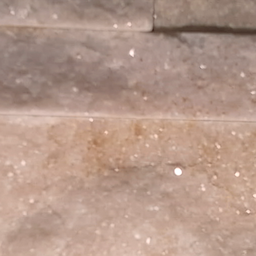}\\

\end{tabularx}

%% file: img/fig_tabular_max_material_54/tabular.tex
\tiny
\begin{tabularx}{\linewidth}{@{}X@{} @{}c@{} @{}c@{} @{}c@{} @{}c@{} @{}c@{} }
& Normal & Diffuse & Roughness & Specular (x10) & Novel light renders \\
\rotatebox{90}{\hspace{3.7mm}Ours} & 
\includegraphics[width=\imgw\linewidth]{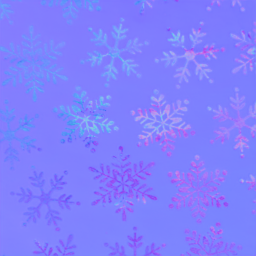} & 
\includegraphics[width=\imgw\linewidth]{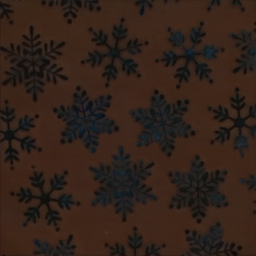} & 
\includegraphics[width=\imgw\linewidth]{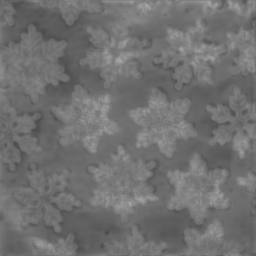} & 
\includegraphics[width=\imgw\linewidth]{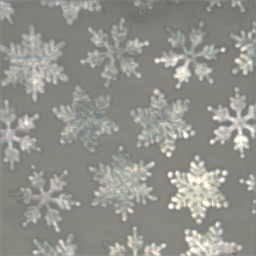} & 
\includegraphics[width=\imgw\linewidth]{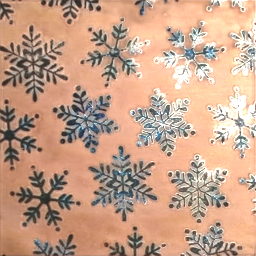}
\includegraphics[width=\imgw\linewidth]{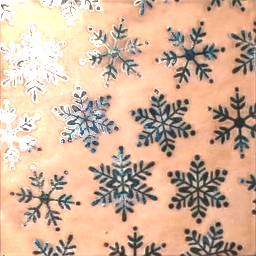}
\includegraphics[width=\imgw\linewidth]{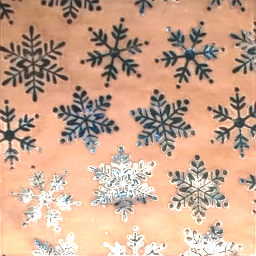}\\
\rotatebox{90}{\hspace{1.8mm}OursFine} & 
\includegraphics[width=\imgw\linewidth]{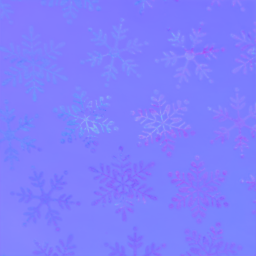} & 
\includegraphics[width=\imgw\linewidth]{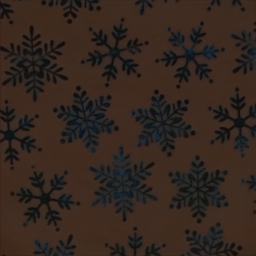} & 
\includegraphics[width=\imgw\linewidth]{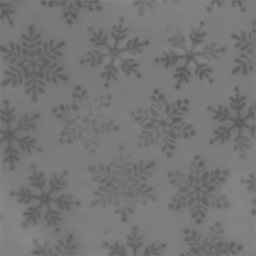} & 
\includegraphics[width=\imgw\linewidth]{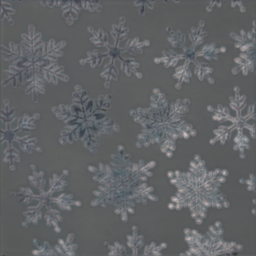} & 
\includegraphics[width=\imgw\linewidth]{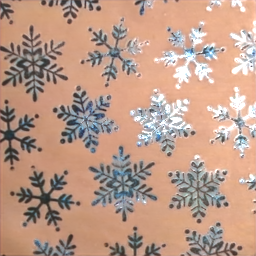}
\includegraphics[width=\imgw\linewidth]{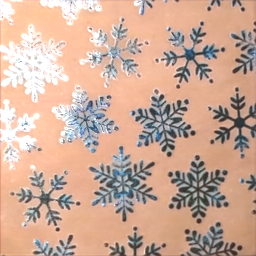}
\includegraphics[width=\imgw\linewidth]{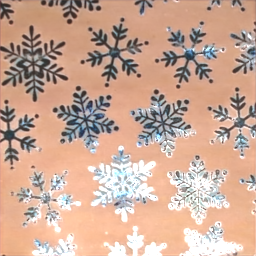}\\
\rotatebox{90}{\hspace{1.8mm}OursOptim} & 
\includegraphics[width=\imgw\linewidth]{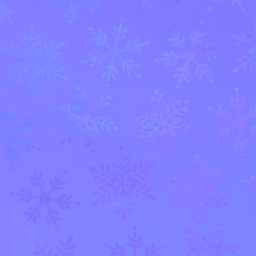} & 
\includegraphics[width=\imgw\linewidth]{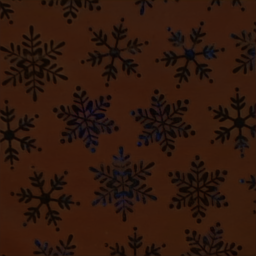} & 
\includegraphics[width=\imgw\linewidth]{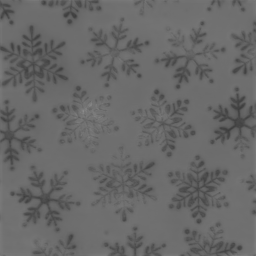} & 
\includegraphics[width=\imgw\linewidth]{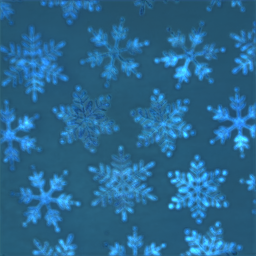} & 
\includegraphics[width=\imgw\linewidth]{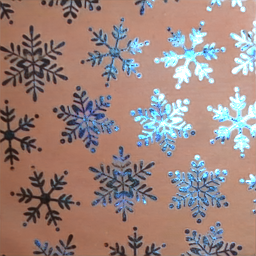}
\includegraphics[width=\imgw\linewidth]{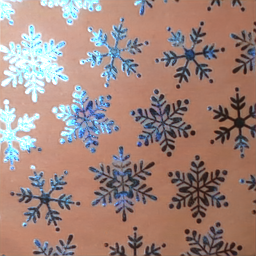}
\includegraphics[width=\imgw\linewidth]{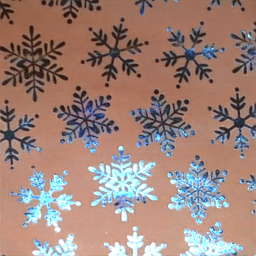}\\
\rotatebox{90}{\hspace{1.8mm}Reference} & 
 & 
 & 
 & 
 & 
\includegraphics[width=\imgw\linewidth]{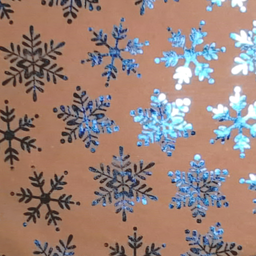}
\includegraphics[width=\imgw\linewidth]{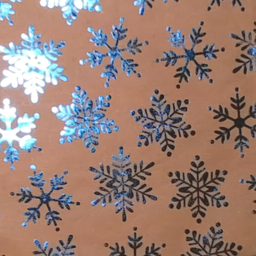}
\includegraphics[width=\imgw\linewidth]{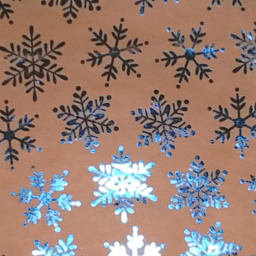}\\

\end{tabularx}

%% file: img/fig_tabular_gao_material_33/tabular.tex
\tiny
\begin{tabularx}{\linewidth}{@{}X@{} @{}c@{} @{}c@{} @{}c@{} @{}c@{} @{}c@{} }
& Normal & Diffuse & Roughness & Specular (x10) & Novel light renders \\
\rotatebox{90}{\hspace{1.1mm}Deschaintre} & 
\includegraphics[width=\imgw\linewidth]{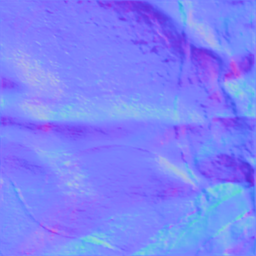} & 
\includegraphics[width=\imgw\linewidth]{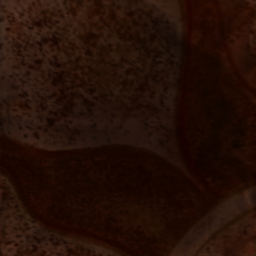} & 
\includegraphics[width=\imgw\linewidth]{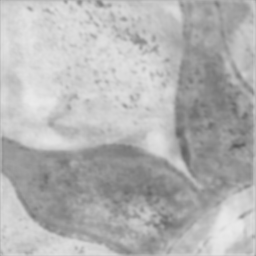} & 
\includegraphics[width=\imgw\linewidth]{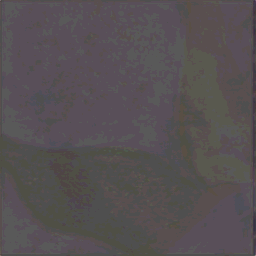} & 
\includegraphics[width=\imgw\linewidth]{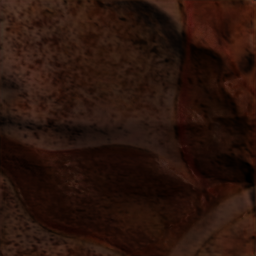}
\includegraphics[width=\imgw\linewidth]{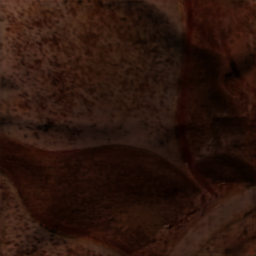}
\includegraphics[width=\imgw\linewidth]{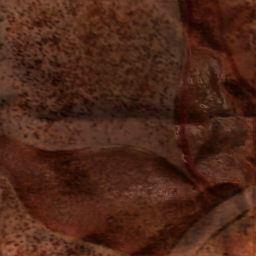}\\
\rotatebox{90}{\hspace{4.0mm}Gao} & 
\includegraphics[width=\imgw\linewidth]{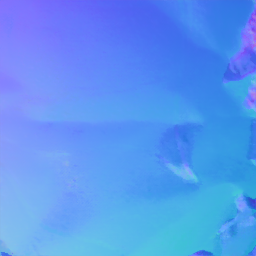} & 
\includegraphics[width=\imgw\linewidth]{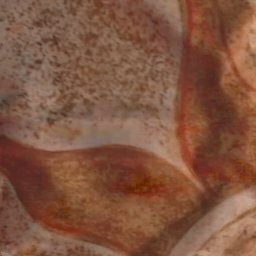} & 
\includegraphics[width=\imgw\linewidth]{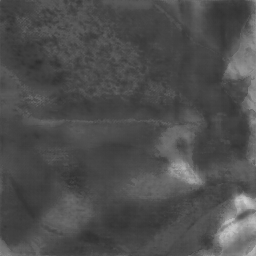} & 
\includegraphics[width=\imgw\linewidth]{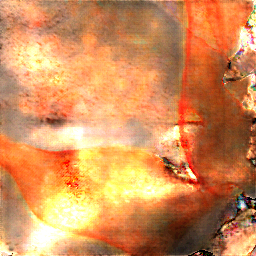} & 
\includegraphics[width=\imgw\linewidth]{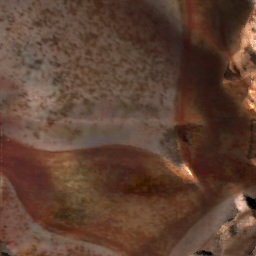}
\includegraphics[width=\imgw\linewidth]{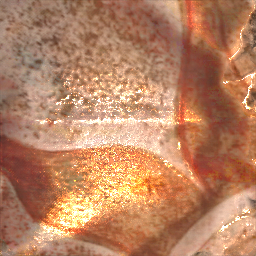}
\includegraphics[width=\imgw\linewidth]{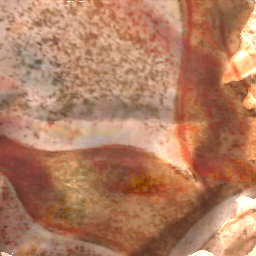}\\
\rotatebox{90}{\hspace{3.7mm}Ours} & 
\includegraphics[width=\imgw\linewidth]{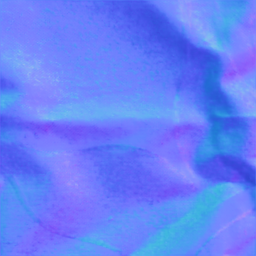} & 
\includegraphics[width=\imgw\linewidth]{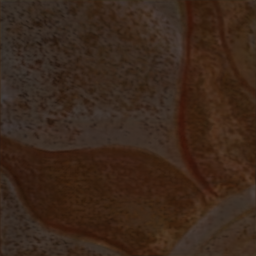} & 
\includegraphics[width=\imgw\linewidth]{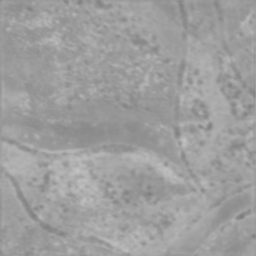} & 
\includegraphics[width=\imgw\linewidth]{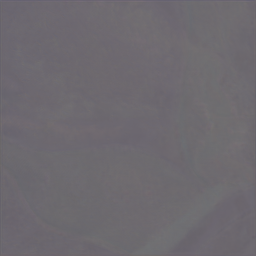} & 
\includegraphics[width=\imgw\linewidth]{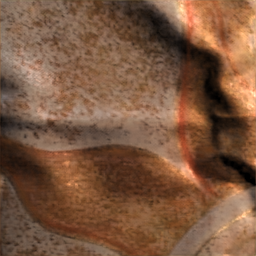}
\includegraphics[width=\imgw\linewidth]{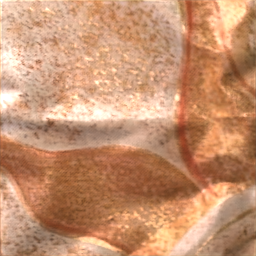}
\includegraphics[width=\imgw\linewidth]{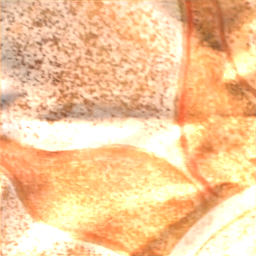}\\
\rotatebox{90}{\hspace{1.8mm}OursOptim} & 
\includegraphics[width=\imgw\linewidth]{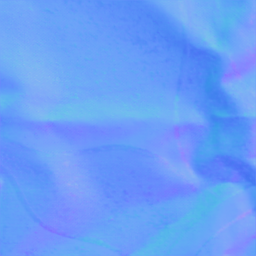} & 
\includegraphics[width=\imgw\linewidth]{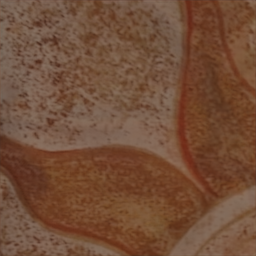} & 
\includegraphics[width=\imgw\linewidth]{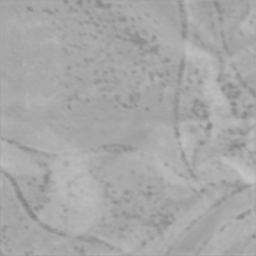} & 
\includegraphics[width=\imgw\linewidth]{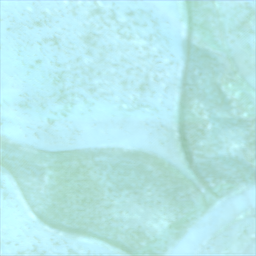} & 
\includegraphics[width=\imgw\linewidth]{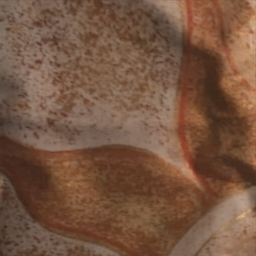}
\includegraphics[width=\imgw\linewidth]{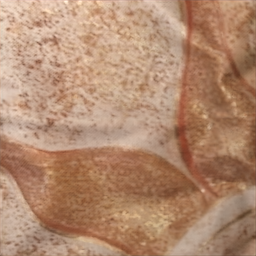}
\includegraphics[width=\imgw\linewidth]{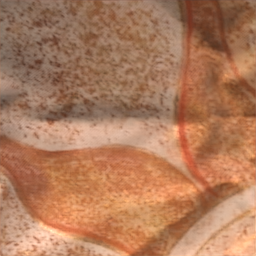}\\
\rotatebox{90}{\hspace{1.8mm}Reference} & 
 & 
 & 
 & 
 & 
\includegraphics[width=\imgw\linewidth]{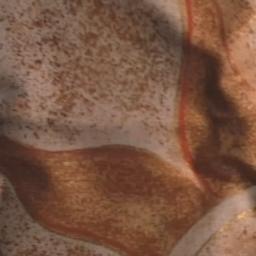}
\includegraphics[width=\imgw\linewidth]{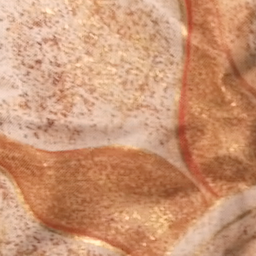}
\includegraphics[width=\imgw\linewidth]{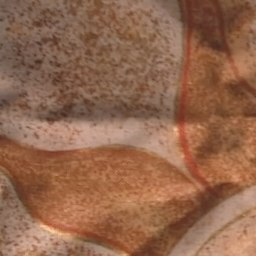}\\

\end{tabularx}

%% file: img/fig_tabular_gao_material_6/tabular.tex
\tiny
\begin{tabularx}{\linewidth}{@{}X@{} @{}c@{} @{}c@{} @{}c@{} @{}c@{} @{}c@{} }
& Normal & Diffuse & Roughness & Specular (x10) & Novel light renders \\
\rotatebox{90}{\hspace{1.1mm}Deschaintre} & 
\includegraphics[width=\imgw\linewidth]{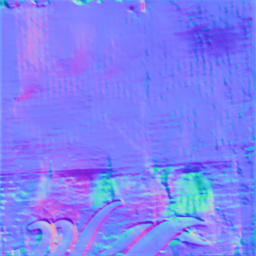} & 
\includegraphics[width=\imgw\linewidth]{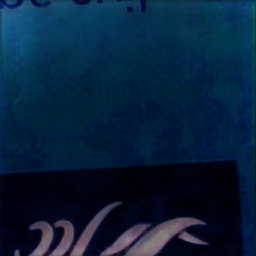} & 
\includegraphics[width=\imgw\linewidth]{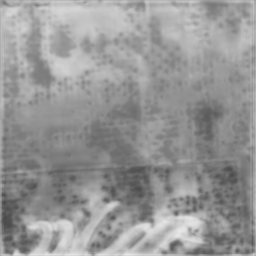} & 
\includegraphics[width=\imgw\linewidth]{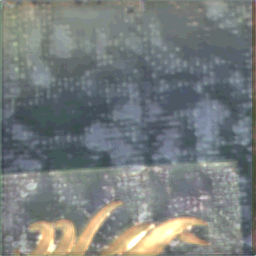} & 
\includegraphics[width=\imgw\linewidth]{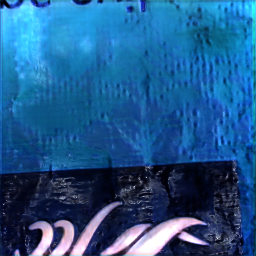}
\includegraphics[width=\imgw\linewidth]{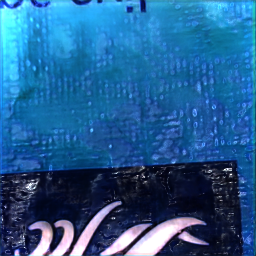}
\includegraphics[width=\imgw\linewidth]{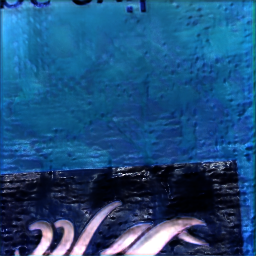}\\
\rotatebox{90}{\hspace{4.0mm}Gao} & 
\includegraphics[width=\imgw\linewidth]{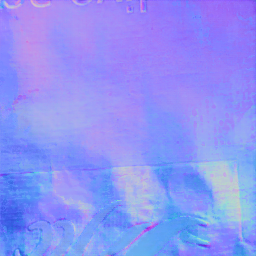} & 
\includegraphics[width=\imgw\linewidth]{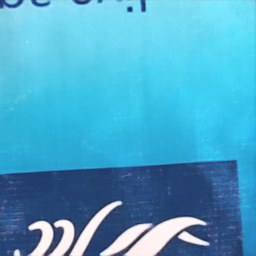} & 
\includegraphics[width=\imgw\linewidth]{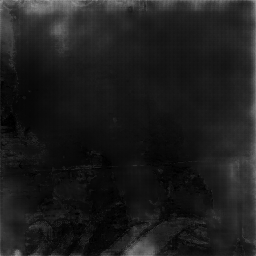} & 
\includegraphics[width=\imgw\linewidth]{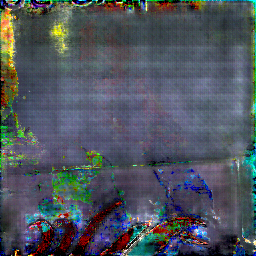} & 
\includegraphics[width=\imgw\linewidth]{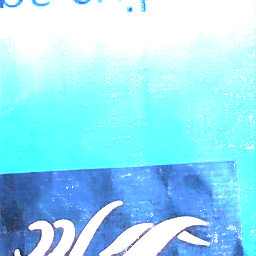}
\includegraphics[width=\imgw\linewidth]{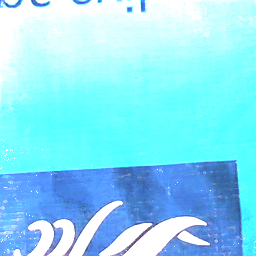}
\includegraphics[width=\imgw\linewidth]{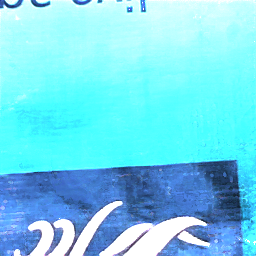}\\
\rotatebox{90}{\hspace{3.7mm}Ours} & 
\includegraphics[width=\imgw\linewidth]{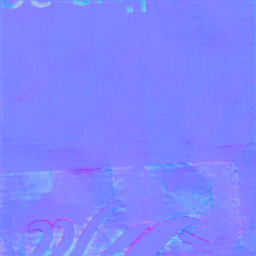} & 
\includegraphics[width=\imgw\linewidth]{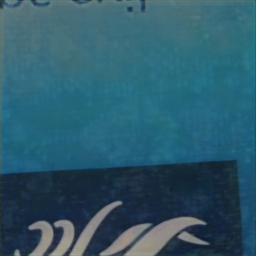} & 
\includegraphics[width=\imgw\linewidth]{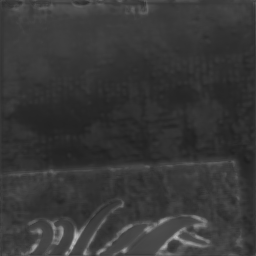} & 
\includegraphics[width=\imgw\linewidth]{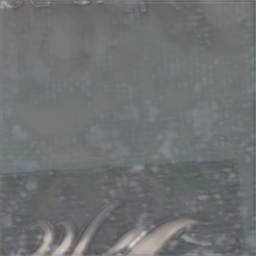} & 
\includegraphics[width=\imgw\linewidth]{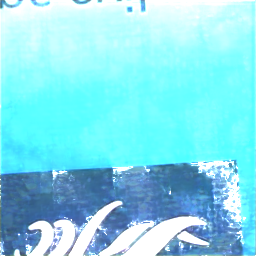}
\includegraphics[width=\imgw\linewidth]{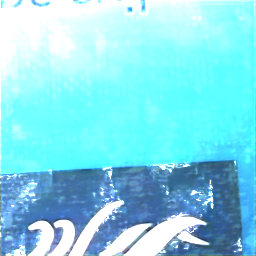}
\includegraphics[width=\imgw\linewidth]{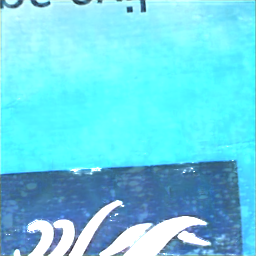}\\
\rotatebox{90}{\hspace{1.8mm}OursOptim} & 
\includegraphics[width=\imgw\linewidth]{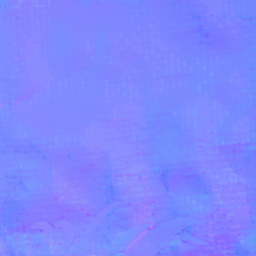} & 
\includegraphics[width=\imgw\linewidth]{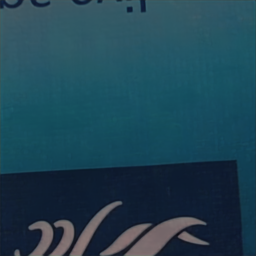} & 
\includegraphics[width=\imgw\linewidth]{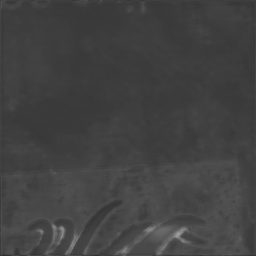} & 
\includegraphics[width=\imgw\linewidth]{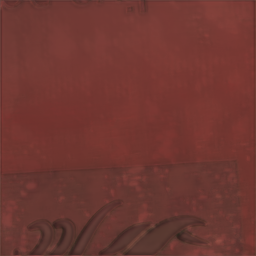} & 
\includegraphics[width=\imgw\linewidth]{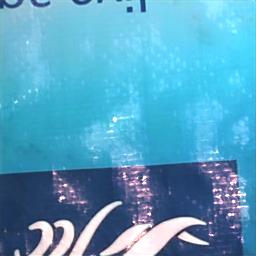}
\includegraphics[width=\imgw\linewidth]{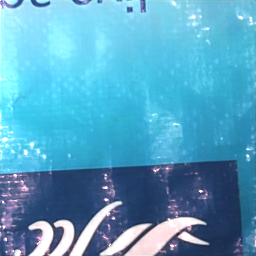}
\includegraphics[width=\imgw\linewidth]{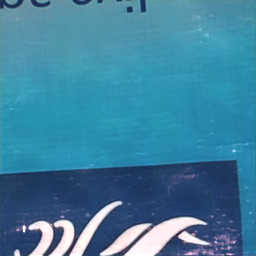}\\
\rotatebox{90}{\hspace{1.8mm}Reference} & 
 & 
 & 
 & 
 & 
\includegraphics[width=\imgw\linewidth]{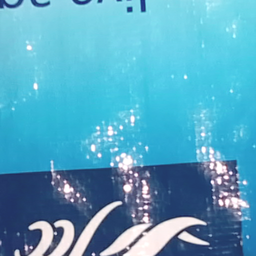}
\includegraphics[width=\imgw\linewidth]{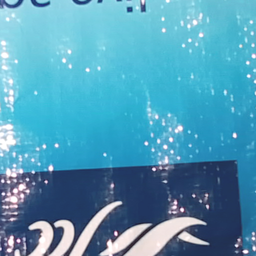}
\includegraphics[width=\imgw\linewidth]{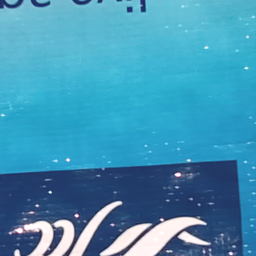}\\

\end{tabularx}

%% file: content/discussion.tex
\section{Discussion}

\begin{figure}[t]
\centering
    \def\imgw{0.138}
\include{img/fig_tabular_limitations_material_25/tabular}
\caption{Example of limitations. Two shiny mugs are placed next to each other, generating reflections and cast shadows which are not modeled by our image formation model. Both the model trained on synthetic data and finetuned on real data do not approximate the SVBRDF accurately. While the per-material optimization significantly improves the results (notably on the normal map), the novel light renders do not properly capture these effects.}
\label{fig:limitations}
\end{figure}

In this work, we demonstrate through extensive quantitative experiments that SVBRDF estimation techniques trained solely on synthetic data do not generalize properly when evaluated on real images. To this end, we capture a novel dataset, captured with a novel multi-light capture system, and composed of 80 real surfaces at various scales, yielding a total of 460 material samples over 5,000 images. Bridging the gap between synthetic and real, our novel dynamic architecture offers state-of-the-art SVBRDF recovery on real-world images. 

Despite the improvements achieved in high-fidelity material recovery using a simple capture system, our method still suffers from some limitations, some of which are shared with previous work. Higher-order light interactions such as reflections, interreflections, and cast shadows are not taken into account in our image formation model. Some failure cases where these assumptions are not satisfied are shown in fig.~\ref{fig:limitations}. While the per-material optimization yields plausible results, visible artifacts are still present in the results. Additionally, our proposed method focuses primarily on material estimation. We hope our work will inspire future developments on material estimation in the wild and be extended to related tasks such as lighting estimation.


%% file: img/fig_tabular_limitations_material_25/tabular.tex
\tiny
\begin{tabularx}{\linewidth}{@{}X@{} @{}c@{} @{}c@{} @{}c@{} @{}c@{} @{}c@{} }
& Normal & Diffuse & Roughness & Specular (x10) & Novel light renders \\
\rotatebox{90}{\hspace{3.7mm}Ours} & 
\includegraphics[width=\imgw\linewidth]{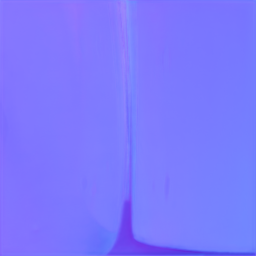} & 
\includegraphics[width=\imgw\linewidth]{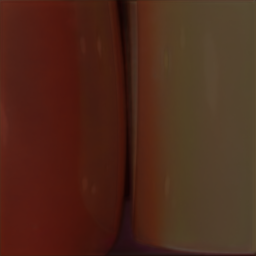} & 
\includegraphics[width=\imgw\linewidth]{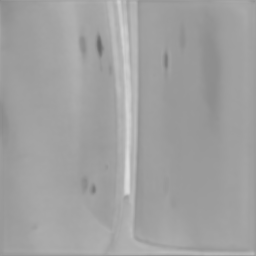} & 
\includegraphics[width=\imgw\linewidth]{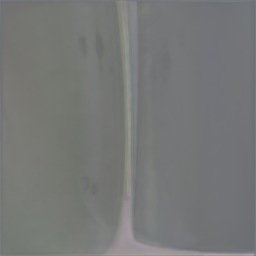} & 
\includegraphics[width=\imgw\linewidth]{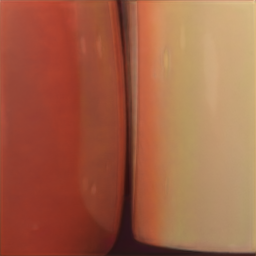}
\includegraphics[width=\imgw\linewidth]{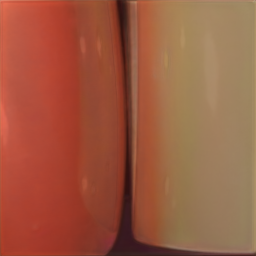}
\includegraphics[width=\imgw\linewidth]{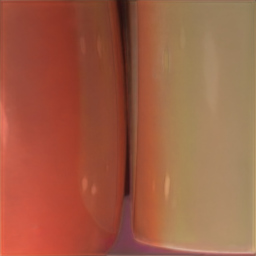}\\
\rotatebox{90}{\hspace{1.8mm}OursFine} & 
\includegraphics[width=\imgw\linewidth]{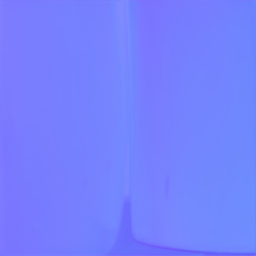} & 
\includegraphics[width=\imgw\linewidth]{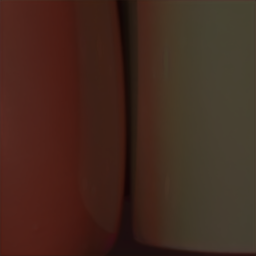} & 
\includegraphics[width=\imgw\linewidth]{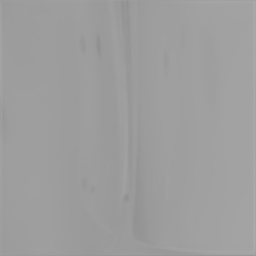} & 
\includegraphics[width=\imgw\linewidth]{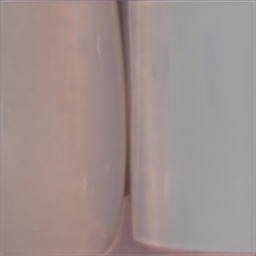} & 
\includegraphics[width=\imgw\linewidth]{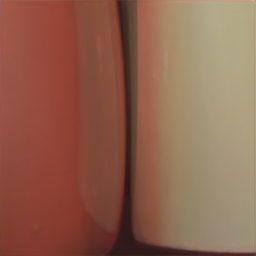}
\includegraphics[width=\imgw\linewidth]{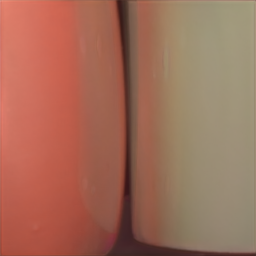}
\includegraphics[width=\imgw\linewidth]{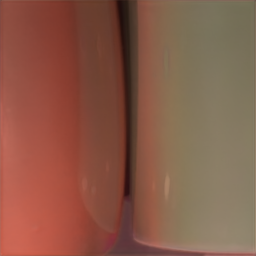}\\
\rotatebox{90}{\hspace{1.8mm}OursOptim} & 
\includegraphics[width=\imgw\linewidth]{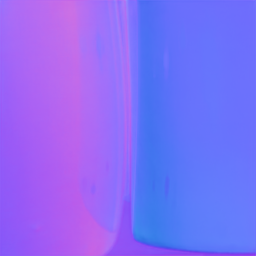} & 
\includegraphics[width=\imgw\linewidth]{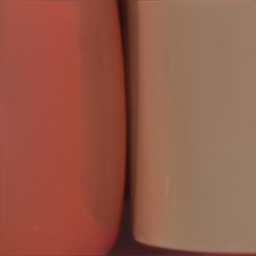} & 
\includegraphics[width=\imgw\linewidth]{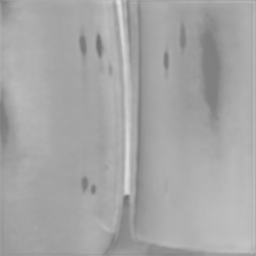} & 
\includegraphics[width=\imgw\linewidth]{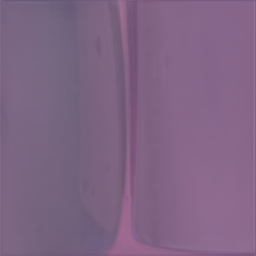} & 
\includegraphics[width=\imgw\linewidth]{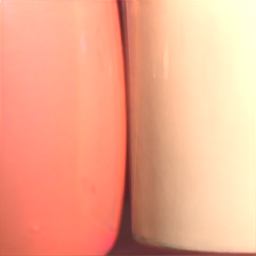}
\includegraphics[width=\imgw\linewidth]{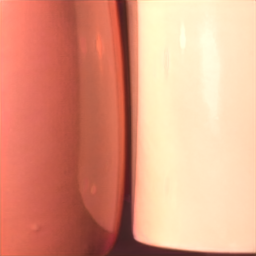}
\includegraphics[width=\imgw\linewidth]{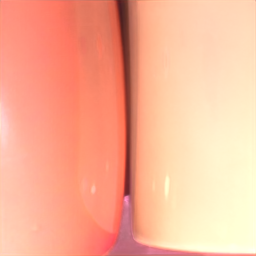}\\
\rotatebox{90}{\hspace{1.8mm}Reference} & 
 & 
 & 
 & 
 & 
\includegraphics[width=\imgw\linewidth]{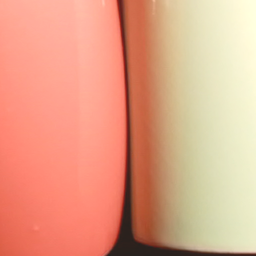}
\includegraphics[width=\imgw\linewidth]{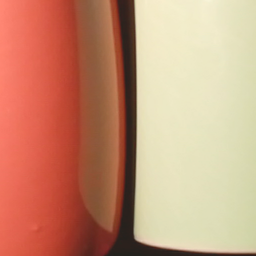}
\includegraphics[width=\imgw\linewidth]{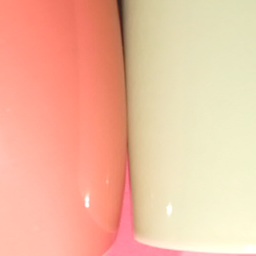}\\

\end{tabularx}